\documentclass[lettersize,journal]{IEEEtran}
\usepackage{amsmath,amsfonts}
\usepackage{algorithm}
\usepackage{algorithmicx}
\usepackage{algpseudocode}
\usepackage{array}
\usepackage[caption=false]{subfig}
\usepackage{textcomp}
\usepackage{stfloats}
\usepackage{url}
\usepackage{multirow}
\usepackage{booktabs}
\usepackage{pifont}
\usepackage{threeparttable}
\usepackage{verbatim}
\usepackage{graphicx}
\usepackage{cite}
\usepackage{epstopdf}
\newtheorem{myDef}{Definition}
\usepackage[normalem]{ulem}
\usepackage{setspace}

\useunder{\uline}{\ul}{}
\def\smallerspacecaption{\vspace{-2mm}}
\begin{document}

\title{Coupled Attention Networks for Multivariate Time Series Anomaly Detection}

\author{Feng Xia,~\IEEEmembership{Senior Member,~IEEE}, Xin Chen, Shuo Yu,~\IEEEmembership{Member,~IEEE}, Mingliang Hou, Mujie Liu, \\and Linlin You,~\IEEEmembership{Member,~IEEE}
\thanks{Corresponding author: Shuo Yu.}
\thanks{This work is partially supported by National Natural Science Foundation of China under Grant No. 62102060 and the Fundamental Research Funds for the Central Universities under Grant No. DUT22RC(3)060.}}

\markboth{}%
{Xia \MakeLowercase{\textit{et al.}}: Coupled Attention Networks for Multivariate Time Series Anomaly Detection}

\IEEEpubid{}

\maketitle

\begin{abstract}
Multivariate time series anomaly detection (MTAD) plays a vital role in a wide variety of real-world application domains. Over the past few years, MTAD has attracted rapidly increasing attention from both academia and industry. Many deep learning and graph learning models have been developed for effective anomaly detection in multivariate time series data, which enable advanced applications such as smart surveillance and risk management with unprecedented capabilities. Nevertheless, MTAD is facing critical challenges deriving from the dependencies among sensors and variables, which often change over time. To address this issue, we propose a coupled attention-based neural network framework (CAN) for anomaly detection in multivariate time series data featuring dynamic variable relationships. We combine adaptive graph learning methods with graph attention to generate a global-local graph that can represent both global correlations and dynamic local correlations among sensors. To capture inter-sensor relationships and temporal dependencies, a convolutional neural network based on the global-local graph is integrated with a temporal self-attention module to construct a coupled attention module. In addition, we develop a multilevel encoder-decoder architecture that accommodates reconstruction and prediction tasks to better characterize multivariate time series data. Extensive experiments on real-world datasets have been conducted to evaluate the performance of the proposed CAN approach, and the results show that CAN significantly outperforms state-of-the-art baselines.
\end{abstract}


\section{Introduction}
\IEEEPARstart{S}{ensors} of various types have been deployed in the real world to perceive the state of entities or systems. The prevalence of smart sensors has enabled advanced infrastructures such as the Internet of Things (IoT) and cyber-physical systems (CPS), which in turn have led to a technological revolution in many domains, including smart cities, automated factories, digital twins, public security, self-driving vehicles, and epidemic disease control. With increased sensing, computing, and communication capabilities, smart sensors continue to generate large amounts of time series data in many real-world systems. Detecting various anomalies in these time series data in an automatic and timely manner is critical to intelligent systems and services~\cite{8926446}. Over these years, a great number of machine learning models and algorithms have been developed to fulfill this demand~\cite{B2021A}. For instance, many researchers~\cite{9347460,Pang2021review} have made use of deep learning and/or graph learning to achieve real-time and accurate anomaly detection.

Despite significant advancements in deep anomaly detection, multivariate time series data, which have become ubiquitous due to the use of a large number of smart sensors, cause unprecedented challenges for anomaly detection~\cite{RuizFLMB21,9525836,kong2021}. In systems that feature multivariate time series data, traditional anomaly detection methods might become inapplicable due to their inability to address multidimensional data and complex/temporal relationships. As a consequence, multivariate time series anomaly detection (MTAD) has attracted much attention in recent years~\cite{9705079,HuangWWW2022,hundman2018detecting}. In particular, deep learning provides a basis for learning complex multidimensional dependencies in multivariate time series~\cite{TranAD2022,USAD2020}.

Another promising line of research in MTAD is the use of graph learning (i.e., machine learning on graphs)~\cite{Xia2021TAI}. Graph learning is applicable to a variety of tasks (such as node classification, graph classification, link prediction, and clustering) in many domains (such as medical diagnosis, knowledge graphs, drug discovery, computer vision, natural language processing, and recommender systems). In particular, graph neural networks (GNNs) are widely used to capture the dependencies among variables~\cite{9471816}. A popular trend is to represent relationships between variables in the form of a graph~\cite{9666863}. In practice, there is often a lack of a priori knowledge about variable correlations in multivariate time series, and consequently, the corresponding graphs are missing. Some researchers (e.g.,~\cite{GTA}) have used adaptive learning methods to learn static graph structures from data. However, it has been recognized that the relationship in time series data is dynamic transformation~\cite{Li_Zhu_2021}, and using a static graph to represent dynamic variable relationships is impracticable. To address this challenge, graph attention networks (GATs)~\cite{velickovic2018graph,MTADGAN2020} have been used to detect dynamic variable interactions while ignoring global correlations between variables. Although GNNs have demonstrated strong representational capabilities in non-Euclidean space~\cite{xiaCenGCN2022, LinResource2022}, GNNs and their variants lack the ability to comprehensively construct dynamic graphs without a priori knowledge.

\par To address the above knowledge gap in MTAD, this paper proposes a novel coupled attention network (CAN), which takes advantage of graph learning, attention mechanisms, and convolutional neural networks. A global-local graph is established to describe the dynamic correlations between variables in multivariate time series. Specifically, learnable embedding vectors are used to represent variable features, and embeddings can learn global correlations between variables from training data. The global graph represents the correlations in the entire time series. Furthermore, the local correlations are obtained through the attention mechanism, and the local graph represents the relationships between variables in the current input local time series. Therefore, the relationships in the global graph are used as candidate neighbors to filter the local graph. The global-local graph convolution module and the temporal self-attention module are combined to construct the coupled attention layer, which is the primary layer of the model.
This layer has the ability to model complex dependencies among variables and temporal dependencies. To fully utilize time series data, we use a novel multilevel encoder-decoder framework that facilitates reconstruction and prediction. The proposed framework consists of an encoder and two decoders for prediction and reconstruction. To minimize the risk of overfitting, reconstructed models are only used to assist in learning multivariate time series representations during training, while predictive models are ultimately used for anomaly detection.

The contributions of this paper are summarized as follows:
\begin{itemize}
  \item \textbf{Global-local graph convolutional network:} To capture complex and dynamic variable correlations in multivariate time series data, we propose a novel global-local graph convolutional network. This network can represent both global correlations and dynamic local correlations among sensors (or variables).
  \item \textbf{Coupled attention network:} We present a coupled attention network based on the proposed global-local graph convolutional network and the attention mechanism. Specifically, a temporal self-attention module is exploited. The integration of graph learning and attention mechanisms ensures our CAN approach has the ability to model complex dependencies among variables and temporal dependencies in the absence of prior knowledge.
  \item \textbf{Multilevel encoder-decoder framework for anomaly detection:} We propose a novel multilevel encoder-decoder framework to solve the problem of multivariate time series anomaly detection. The framework is a prediction-based anomaly detection method and uses reconstruction tasks to jointly represent time series data.
  \item \textbf{Comprehensive experimental evaluation:} We conducted extensive experiments on three real-world multivariate time series datasets. The experimental results demonstrate the effectiveness and superiority of the proposed approach, CAN, in comparison with state-of-the-art baselines.
\end{itemize}
\par The remainder of the paper is organized as follows: We review related work on time series anomaly detection in Section 2. Section 3 provides relevant preliminaries in which the problem statement is given. The details of the proposed framework CAN are presented in Section 4, and in Section 5, the performance evaluation of CAN is discussed, and the experimental results are analyzed. Finally, we conclude the paper in Section 6.

\section{Related Work}
Time series can be univariate or multivariate. Accordingly, we categorize time series anomaly detection into univariate (time series) anomaly detection and multivariate (time series) anomaly detection. Univariate anomaly detection models each variable independently, while multivariate anomaly detection considers the connection between multiple variables to detect anomalies.

\subsection{Univariate Anomaly Detection}
Before deep learning became popular, various mathematical and statistical models were developed to analyze time series data, most of which have already been used for anomaly detection. For example, the autoregressive integrated moving average (ARIMA)~\cite{box1970distribution} is used to forecast future states based on past states. The discrepancy between the forecasted value and the ground truth is used to deduce anomalies. In~\cite{MARKOU20032481}, the mathematical analysis of time series data generates a statistical model by calculating statistical measures, such as the mean, variance, and quantile. The statistical model can detect the tested data to determine whether it belongs to the normal boundary. Recently, machine learning and deep learning methods have achieved remarkable improvements in time series anomaly detection~\cite{ERHAN202164,Pang2021review,B2021A}. In particular, autoencoders (AEs)~\cite{AE} and variational autoencoders (VAEs)~\cite{kingma2013auto} are widely used as reconstruction models that employ reconstruction errors as anomaly scores. For example, DAGMM~\cite{zong2018deep} combines Gaussian mixture models and AEs to obtain reconstruction models for anomaly detection. In addition, long short-term memory (LSTM) has also been frequently utilized in time series modeling, and researchers have introduced LSTM-based autoencoders for reconstruction-based anomaly detection~\cite{lstmEncoder}.

\subsection{Multivariate Time Series Anomaly Detection}
In the context of MTAD, there are often correlations between variables in the same entity or system in the real world. Many researchers have started using multivariate time series data as input to improve anomaly detection accuracy~\cite{HuangWWW2022,TranAD2022,USAD2020,MTADGAN2020}. MTAD methods based on deep learning can be roughly classified into two categories: reconstruction-based and prediction-based.

Reconstruction-based methods learn low-dimensional representations of multivariate time series, reconstruct normal values, and detect anomalies based on reconstruction errors. For example, LSTM-VAE~\cite{LSTM-VAE} combines LSTM with a VAE to fuse signals and reconstruct the expected distribution. The LSTM-based encoder projects multivariate observations into a latent space. The decoder then reconstructs the expected distribution of the multivariate inputs. OmniAnomaly~\cite{OmniAnomaly} uses a stochastic recurrent neural network to avoid potential misguidance by uncertain instances. Its core idea is to build robust representations of multivariate data to capture their normal patterns and reconstruct the input data. The above methods reconstruct the input sequence but also reconstruct the constructed features. MSCRED~\cite{Zhang_2019} detects anomalies by reconstructing the signature matrices of multiscale relationships between variables. These methods are good at capturing global data distributions. Prediction-based methods attempt to predict the normal values of indicators based on historical data and detect anomalies based on prediction errors. For example, LSTM-NDT~\cite{hundman2018detecting} uses an LSTM-based prediction approach with unsupervised nonparametric dynamic thresholding for anomaly detection. These methods are specialized for feature engineering in the prediction of the next timestamp. Reconstruction-based and prediction-based methods have their own advantages, but few methods consider joint reconstruction and prediction tasks to simultaneously characterize multivariate time series data~\cite{han2022kdd,DVGCRN}. In addition to reconstruction-based and prediction-based anomaly detection, many novel anomaly detection methods have emerged, including base classification anomaly detection~\cite{shenNEURIPS2020} and contrast learning-based anomaly detection~\cite{jiao2022}.

\begin{table}[tbp]
  \caption{Review of existing solutions.}
  \centering
  \label{relatedWork}
  \smallerspacecaption
  \small
  \begin{threeparttable}
    \setlength{\tabcolsep}{3mm}{
    \begin{tabular}{@{}cccccc@{}}
    \toprule
    & Related work & C1\tnote{1} & C2\tnote{1} & C3\tnote{2} & C4\tnote{2} \\ \midrule
                & ARIMA       & \ding{56} & \ding{52} & - & - \\
                & KNN         & \ding{56} & \ding{56} & - & - \\
    Univariate   & PCA         & \ding{52} & \ding{56} & - & - \\
                & AE          & \ding{52} & \ding{56} & - & - \\
                & IF          & \ding{56} & \ding{56} & - & - \\
                & DAGMM       & \ding{52} & \ding{56} & - & - \\ \midrule
                & LSTM-VAE    & \ding{52} & \ding{56} & \ding{56} & \ding{56} \\
                & OmniAnomaly & \ding{52} & \ding{56} & \ding{56} & \ding{56} \\
    Multivariate & MTAD-GAN    & \ding{52} & \ding{52} & \ding{56} & \ding{52} \\
                & GDN         & \ding{56} & \ding{52} & \ding{52} & \ding{52} \\
                & GTA         & \ding{56} & \ding{52} & \ding{52} & \ding{56} \\
                & DVGCRN        & \ding{52} & \ding{52} & \ding{52} & \ding{52} \\
                \bottomrule
    \end{tabular}
    }
  \begin{tablenotes}
    \footnotesize
    \item[1] C1 and C2 represent Reconstruction and Prediction tasks, respectively.
    \item[2] C3 and C4 represent Global and Local relationships, respectively.
  \end{tablenotes}
  \end{threeparttable}
\end{table}

\par

 Although these methods use multivariate time series as input, none of them explicitly learn or demonstrate the relationship between different time series. This relationship is critical for the diagnosis of anomalies. Graph-structured data have been widely utilized to describe variable relationships in a variety of domains, such as smart cities, transportation, and finance. In addition, graph convolutional networks (GCNs)~\cite{kipf2017semi} and GATs~\cite{velickovic2018graph} are widely used in various fields. For instance, the GDN~\cite{GDN} depends on sensor embedding to flexibly capture the characteristics of each sensor and learns the relationships between sensors by cosine similarity. MTAD-GAN~\cite{MTADGAN2020} concatenates two graph attention layers in parallel to learn the complex dependencies in temporal and feature dimensions. GTA~\cite{GTA} uses a transformer-based architecture to model temporal correlation and proposes a self-learning graph structure to capture bidirectional links between sensors. DVGCRN~\cite{DVGCRN} designs adaptive variational graph convolutional recurrent networks to model spatial and temporal fine-grained correlations and capture multilevel information at different layers.

 \par Table~\ref{relatedWork} provides an evaluation of the relevant models, where KNN~\cite{KNN}, PCA~\cite{LI201463}, and IF~\cite{liu2008isolation} stand for k-nearest neighbors, principal component analysis, and isolated forests, respectively. Almost all existing methods use only one task to train the model, and rarely do they combine both tasks to train the model. However, the reconstruction and prediction tasks actually have the potential to complement each other. In addition, although many researchers use GNNs to capture the correlation between variables, constructing the graph and representing the dynamics of the graph for MTAD has not been solved. Global relationships that depend on the characteristics of variables and local relationships that change over time are not considered simultaneously. In this paper, we propose an effective approach to address these challenges.

\section{Preliminaries}
The main focus of the paper is on MTAD at the entity level. This section introduces the definitions and then formalizes the MTAD problem.
\begin{table}[tbp]
  \caption{Notations.}
  \smallerspacecaption
  \setlength{\tabcolsep}{1mm}{
  \footnotesize
  \begin{tabular}{@{}cl@{}}
  \toprule
  Notations & Description \\ \midrule
  $\mathcal{A}$ & The entity or system being detected \\
  $\mathcal{X}$ & The original training data \\
  $\mathcal{W}$ & The windowed series of training data $\mathcal{X}$ \\
  $\boldsymbol{X}$ & A multivariate time series of entity $\mathcal{A}$ \\
  $\boldsymbol{y}$ & The vector containing the results of anomaly detection \\
  $\boldsymbol{Y}$ & The result of prediction \\
  $\mathcal{F}$ & A forecast function \\
  $\mathcal{C} $ & An anomaly detection function \\
  $\boldsymbol{H}$ & Hidden states \\
  $\boldsymbol{A}^g$ & Global adjacency matrix \\
  $\boldsymbol{A}^l$ & Local adjacency matrix \\
  $\boldsymbol{c}$ & A column vector for calculating the attention coefficient \\
  $K^m$ & The number of relations obtained by filtering \\
  $\boldsymbol{X}_0$ & A placeholder in temporal data \\
  $\boldsymbol{X}_e$ & Sequence embedding \\
  $\mathcal{L}$ & Loss function \\
  $\operatorname{Err}_{n}(t)$ & Prediction deviation of sensor n at timestamp $t$ \\
  $K^s$ & The number of sensors obtained by filtering \\
  $s_t$ & The anomaly score at timestamp $t$ \\ \bottomrule
  \end{tabular}
  }
  \label{notation}
\end{table}

\subsection{Notations and Definitions}
Regarding notations, lowercase letters (e.g., $a$) represent variables, while uppercase letters (e.g., $A$) represent constants. Moreover, letters in bold (e.g., $\boldsymbol{a}$, $\boldsymbol{A}$) represent vectors or matrices. For the sake of readability, the notations used throughout the paper are listed in Table ~\ref{notation}.

\begin{myDef}
	\textbf{Multivariate Time Series (MTS):}
	A time series is a collection of observations recorded at equal-space timestamps. In this paper, assuming that an entity or system $\mathcal{A}$ contains multiple features or sensors, the characteristics of entity $\mathcal{A}$ over a period of time can be expressed as MTS $\boldsymbol{X}\in \mathbb{R}^{N\times L}$, where $L$ is the sequence length and $N$ is the number of features or sensors. Moreover, $\boldsymbol{X}_{n,:} \in \mathbb{R}^L$ represents the $n^{th}$ feature ($n \in \{0, 1, \ldots, N-1\}$) of $\mathcal{A}$ throughout all timestamps, and $\boldsymbol{X}_{:,t} \in \mathbb{R}^N$ represents all features of $\mathcal{A}$ at the $t^{th}$ timestamp ($t\in \{0, 1, \ldots, L-1\}$).
\end{myDef}
\begin{myDef}
	\textbf{Sensor Graph (SG):} The SG is represented as $\mathcal{G} = \{\mathcal{V} , \boldsymbol{A}\} $, where $\mathcal{V}=\{0, 1, \ldots, N-1\}$ is a set of nodes (in terms of sensors or their features) and $\boldsymbol{A}\in \mathbb{R}^{N\times N}$ is the adjacency matrix of N nodes in which a cell $a_{i,j}$ is used to record the edge weight between the $i^{th}$ and $j^{th}$ nodes.
\end{myDef}

\begin{figure*}
	\centering
	\includegraphics[width=17cm]{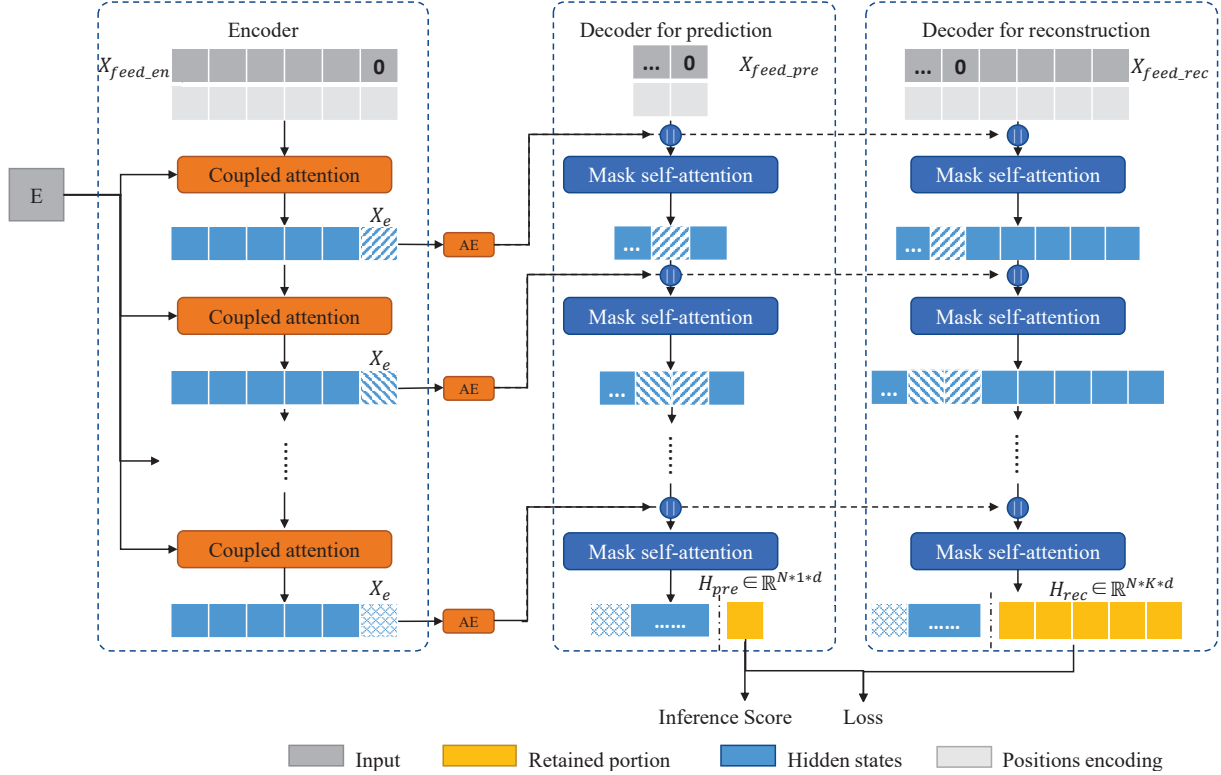}\\
	\makeatletter\def\@captype{figure}\makeatother
	\caption{ Overall CAN framework with one encoder and two decoders. The rounded rectangles represent neural networks, and the sharp rectangles represent tensors. The normal linear layers are hidden. The last layer of the two decoders intercepts the tensor to obtain a tensor of target length $1$ and $K$. During training, the model is jointly trained by the outputs of the two decoders. Only the predicted decoder operates to calculate the anomaly score when detecting anomalies.}
  \label{model}
  \smallerspacecaption
\end{figure*}

\subsection{Problem Statement}
In general, MTAD determines whether an entity or system is anomalous at the $t^{th}$ timestamp based on the corresponding observation $\boldsymbol{X}_{:,t}$:
\begin{equation}
  \boldsymbol{y}_t = \mathcal{C}(\boldsymbol{X}_{:,t}),
  \label{problem1}
\end{equation}
where $\boldsymbol{y}_t$ is a Boolean value indicating the state of the system at the $t^{th}$ timestamp (if $\boldsymbol{y}_t=1$, the system is anomalous at the $t^{th}$ timestamp; otherwise not) and $\mathcal{C}$ denotes the anomaly detection function.

Moreover, as historical data are beneficial in terms of analyzing the current state of the system, a period of historical data can be utilized to support MTAD. Accordingly, Equation~\eqref{problem1} can be rewritten as Eq. \ref{problem2}:
\begin{equation}
  \boldsymbol{y}_t = \mathcal{C}(\boldsymbol{X}_{:,t-K:t-1}, \boldsymbol{X}_{:,t}),
  \label{problem2}
\end{equation}
where $K$ represents the length of the historical data and $\boldsymbol{X}_{:,t-K:t-1}$ is the observation of $K$ historical timestamps. By comparing $\boldsymbol{X}_{:,t}$ with the patterns presented by $\boldsymbol{X}_{:,t-K:t-1}$, the detection function $\mathcal{C}$ generates the detection result.

To compare the current observation with the historical observations for MTAD, a prediction-based approach is studied in this paper; it determines the anomaly based on prediction bias. Specifically, following the unsupervised anomaly detection formulation~\cite{GDN,GTA,OmniAnomaly}, the prediction model is trained based on normal data $\mathcal{X}$ and used to make predictions about data containing anomalies $\mathcal{X}'$. According to the definition of $K$, $\mathcal{X}$ and $\mathcal{X}'$ can be transformed into two windowed series $\mathcal{W} = \{[\boldsymbol{X}_{:,0:K-1}, \boldsymbol{X}_{:,K}], [\boldsymbol{X}_{:,1:K}, \boldsymbol{X}_{:,K+1}], \ldots\}$ and $\mathcal{W}'$ with the same sliding window (i.e., $K+1$). Furthermore, during the training process, given a sample $[\boldsymbol{X}_{:,t-K:t-1}, \boldsymbol{X}_{:,t}]$ from $\mathcal{W}$, $\boldsymbol{X}_{:,t-K:t-1}$ are used to predict $\boldsymbol{X}_{:,t}$, which can be expressed in Eq. \ref{eqtraining} as follows:
\begin{equation}
\label{eqtraining}
  \boldsymbol{Y}_{:, t}=\mathcal{F}_{\boldsymbol{\theta}}(\boldsymbol{X}_{:, t-K:t-1}),
\end{equation}
where $\boldsymbol{Y}_{:, t}$ denotes the prediction result at the $t^{th}$ timestamp, and $\theta$ denotes the parameters of the predicting model. Note that the training goal is to find a function $\mathcal{F}_{\boldsymbol{\theta}}$ to minimize the gap between $\boldsymbol{Y}_{:, t}$ and $\boldsymbol{X}_{:,t}$.

During the testing/detecting process, the same function $\mathcal{F}_{\boldsymbol{\theta}}$ is used to predict $\boldsymbol{Y}'_{:,t}$ based on $\boldsymbol{X}'_{:,t-K:t-1}$. Then, the predicted value and the actual value are passed to $\mathcal{C}$ to calculate the system state $\boldsymbol{y}_t$ according to Eq. \ref{eqpredtodetect} as follows:
\begin{equation}
\label{eqpredtodetect}
  \boldsymbol{y}_t = \mathcal{C} (\mathcal{F}_{\boldsymbol{\theta}}(\boldsymbol{X}'_{:, t-K:t-1}), \boldsymbol{X}'_{:,t}),
\end{equation}

To make the final determination on the anomaly, $\mathcal{C}$ can be implemented with two steps: first, the anomaly score is computed, and then, it is compared with a chosen threshold to obtain the final result.

\section{Methodology}
As shown in Fig.~\ref{model}, a multilevel coupled attention network (CAN) is proposed to support MTAD with the three following components:

\begin{itemize}
  \item An encoder network $Encoder$: The $Encoder$ consists of multiple layers of coupled attention modules to process training samples with position encodings;
	\item A decoder network for prediction $Decoder_{pre}$: The $Decoder_{pre}$ consists of multiple layers of mask self-attention modules to make the prediction based on training samples with position encoding and the representation vectors of $Encoder$;
	\item A decoder network for reconstruction $Decoder_{rec}$: $Decoder_{rec}$ has a structure similar to $Decoder_{pre}$, but it is used to reconstruct the input processed by $Encoder$.
\end{itemize}

These three components make up the CAN framework. In general, the model input is first expanded with a new dimension, and a positional embedding is added in the dimension, similar to other multivariate time series models~\cite{AGCRN}. Second, the encoder represents the input multivariate time series data using layered coupled attention modules and transmits the representation vectors to two decoders. The encoder and two decoders form encoder-decoder structures for prediction and reconstruction, respectively. Third, based on this multilevel structure, the model is trained by minimizing the joint loss of prediction and reconstruction. Finally, the trained model is used on the test set to infer the predictions and detects anomalies based on the predicted and actual values.

\subsection{Coupled Attention Module}
\begin{figure}
	\centering
	\includegraphics[width=8cm]{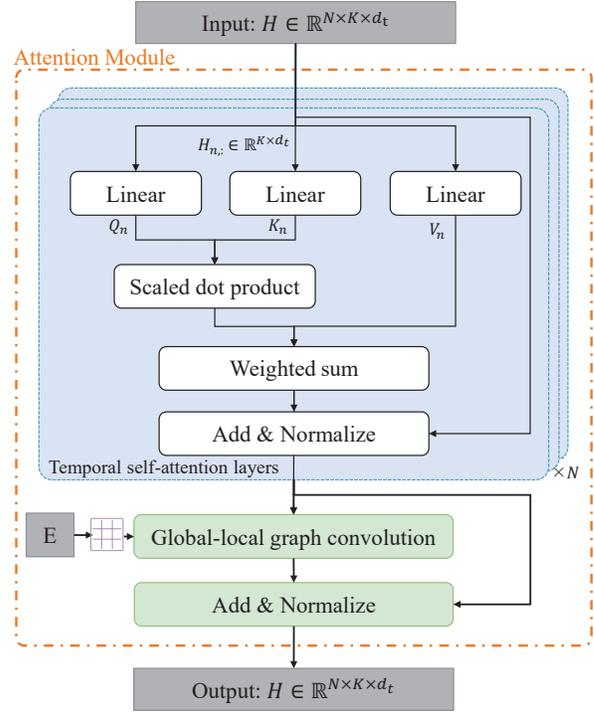}\\
	\makeatletter\def\@captype{figure}\makeatother
	\caption{Coupled attention module consisting of $N$ parallel temporal self-attention layers and a global-local graph convolutional layer.}
  \label{layer}
  \smallerspacecaption
\end{figure}

Based on attention mechanisms, the coupled attention module (CAM) is designed to learn the complex temporal and variable-dependent relationships in multivariate time series data. As shown in Fig.~\ref{layer}, CAM consists of a set of temporal self-attention layers to learn temporal correlations in time series and a global-local graph convolutional layer to learn macro and micro correlations between variables. In addition, the input to CAM is a 3D tensor, denoted as $H\in \mathbb{R}^{N\times K\times d_t}$, with $d_t$ channels for $N$ sensors at $K$ timestamps. Layer normalization and residual connections are applied to solve the gradient disappearance problem and improve the learning performance.

\subsubsection{Temporal Self-attention Layer}
The self-attention mechanism can effectively and efficiently learn temporal dependencies due to its superior capability for sequential data. Regarding MTS data $\boldsymbol{H}\in \mathbb{R}^{N\times K\times d_t}$, $N$ parallel self-attention layers are used to learn the temporal dependence of each sensor separately. Particularly, for the $n^{th}$ sensor, a single layer input is represented by $\boldsymbol{H}_{n}\in \mathbb{R}^{K\times d_t}$.

To compute temporal dependencies, first, the input $\boldsymbol{H}_{n}$ is passed to three full connection layers  separately. Then, $\boldsymbol{H}_{n}$ is projected into three high-dimensional latent subspaces, namely, a query subspace $\boldsymbol{Q_n} \in \mathbb{R}^{K\times d_k}$, a key subspace by $\boldsymbol{K_n} \in \mathbb{R}^{K\times d_k}$ and a value subspace by $\boldsymbol{V_n} \in \mathbb{R}^{K\times d_v}$ according to Eq. \ref{eqtrans}:

\begin{equation}
\label{eqtrans}
  \begin{aligned}
    \boldsymbol{Q}_{n} &=\boldsymbol{H}_{n} \boldsymbol{W}_{q}, \\
    \boldsymbol{K}_{n} &=\boldsymbol{H}_{n} \boldsymbol{W}_{k}, \\
    \boldsymbol{V}_{n} &=\boldsymbol{H}_{n} \boldsymbol{W}_{v},
  \end{aligned}
\end{equation}
where $\boldsymbol{W}_{q} \in \mathbb{R}^{d_t\times d_k}$, $\boldsymbol{W}_{k}\in \mathbb{R}^{d_t\times d_k}$, and $\boldsymbol{W}_{v}\in \mathbb{R}^{d_t\times d_v}$ are learnable parameters.

Second, the scaled dot-product of $\boldsymbol{Q}_{n}$ and $\boldsymbol{K}_{n}$ is used to compute the temporal dependencies, and then, the related results are used to update $\boldsymbol{V}_{n}$ for the layer output. Accordingly, the procedure can be described by Eq. \ref{eqatt}:
\begin{equation}
\label {eqatt}
  \text { Attention }(\boldsymbol{Q}_{n}, \boldsymbol{K}_{n}, \boldsymbol{V}_{n})=\operatorname{Softmax}\left(\frac{\boldsymbol{Q}_{n} (\boldsymbol{K}_{n})^{T}}{\sqrt{d_{k}}}\right) \boldsymbol{V}_{n}.
\end{equation}

Finally, to capture richer information from the three representational subspaces, the multi-head attention mechanism is applied, and it can be expressed by Eq. \ref{eqhead} and Eq. \ref{eqheadcon}:
\begin{equation}
  \label{eqhead}
    \text{Head}_i = \text {Attention}\left(\boldsymbol{H}_n \boldsymbol{W}_{q}^{i}, \boldsymbol{H}_n \boldsymbol{W}_{k}^{i}, \boldsymbol{H}_n \boldsymbol{W}_{v}^{i}\right),
\end{equation}
\begin{equation}
  \label{eqheadcon}
    \text{MultiHead}(\boldsymbol{H}_n) =(\mathop{\Vert}\limits_{i=1}^{h}\text{Head}_i) \boldsymbol{W}_o,
\end{equation}
where $h$ is the number of heads, $\mathop{\Vert}$ denotes the concatenation operation, and $\boldsymbol{W}_o$ contains parameters to be learned.
Both static global correlation and dynamic local correlation between variables exist in multivariate time series. From a global view, the correlation among sensors of the same type can be easily established over a long time series because the curves of the same senor type are similar. However, when the correlation changes, employing local information is critical. For example, in a water treatment system, we can easily establish the connection between two water quality sensors in close proximity. However, the connection may be broken by the closing of a particular gate. To learn about such complex dependencies, we represent them with a global-local graph. The global graph is used to represent static relationships based on sensor similarity, and the local graph is employed to dynamically adjust the global graph based on the local sequence. In other words, the local graph filters out potential neighbors that are first identified by the global graph to establish and formulate a real-time relationship.

We define a learnable embedding $\boldsymbol{e_n}$ representing the inherent characteristics of the $n^{th}$ sensor. We initialize the vector using random initialization (using prior knowledge to initialize the vector when prior knowledge exists). Accordingly, an undirected adjacency matrix $\boldsymbol{A}^g$ representing the relationship among global sensors can be created by calculating the similarity of their embeddings as defined in Eq. \ref{eqadj}:
\begin{equation}
  \boldsymbol{A}^g = Relu(\boldsymbol{E}\times \boldsymbol{E}^T),
  \label{eqadj}
\end{equation}
where $\boldsymbol{E}$ represents the sensor embedding vector matrix (consisting of $\boldsymbol{e_n}$).

The global adjacency matrix $\boldsymbol{A}^g$ represents the general similarity of the sensors, but it cannot measure their time-varying local relationships ~\cite{Li_Zhu_2021}. Inspired by GAT, a local adjacency matrix $\boldsymbol{A}^l$ is created based on an enhanced attention mechanism, which processes the temporal input together with the intrinsic features of the sensor to calculate the attention coefficient $a_{i,j}^l$ as follows:

\begin{equation}
  \boldsymbol{W}\boldsymbol{H}_{(size=(N, K, d_s))} \stackrel{\text { reshape }}{\longrightarrow} \boldsymbol{V}_{(size=(N,K\times d_s))},
  \label{local1}
\end{equation}

\begin{equation}
  e_{i,j}^l =\operatorname{LeakyReLU}\left(\boldsymbol{c}\left([\boldsymbol{v}_{i} \| \boldsymbol{v}_{j}]\right)\right),
\end{equation}
\begin{equation}
  a_{i,j}^l =\frac{\exp \left(e_{i,j}^l\right)}{\sum_{l=1}^{N} \exp \left(e_{i,j}^l\right)},
  \label{local2}
\end{equation}
where $\boldsymbol{W}$ is a weight matrix that transforms the feature dimension, ``$\|$'' represents the concatenation of two nodes' representations, $\boldsymbol{c}\in \mathbb{R}^{2Kd_s}$ is a column vector of learnable parameters, and LeakyReLU denotes a nonlinear activation function. Since $\boldsymbol{A}^l$ depends on the current input, it can reflect the local inter-sensor dependencies.

Based on the synergistic relationship between the two graphs, $\boldsymbol{A}^g$ and $\boldsymbol{A}^l$ are used to design a new graph convolution layer to further model the influence propagation process and update the representation of each node by combining information from neighboring nodes. Specifically, first, the mask adjacency matrix $\boldsymbol{A}^m$ of the neighbor candidates is selected from $\boldsymbol{A}^g$ by filtering the $K^m$ most significant values according to Eq. \ref{eqselect}:

\begin{equation}
\label{eqselect}
  a^{m}_{i,j} = 1 \{i\in \{1, 2, \ldots, N\}, j \in argtopK^m(a_{i, :}^g)\}.
\end{equation}

Second, the information propagation step is defined as follows:

\begin{equation}
  \begin{array}{c}
    \boldsymbol{A}^{gl} = \boldsymbol{A}^{m}{(\boldsymbol{A}^l+\tilde{\boldsymbol{A}}^g)},\\
    \boldsymbol{H}= (\beta \boldsymbol{H}+(1-\beta)\boldsymbol{A}^{gl} \boldsymbol{H})\boldsymbol{W},
  \end{array}
  \label{gcn}
\end{equation}
where $\beta$ denotes a hyperparameter controlling the ratio that retains the original states, $\tilde{\boldsymbol{A}}^g$ is the normalized ${\boldsymbol{A}}^g$, and $\boldsymbol{W}$ represents a learnable parameter matrix.

\subsection{Encoder}
The encoder is the most important part of the whole framework. Regarding the input, the encoder's input is $\boldsymbol{X}_{feed\_en} = \{\boldsymbol{X}, \boldsymbol{X}_0\}$, where $\boldsymbol{X}$ is equivalent to $\boldsymbol{X}_{:, t-K:t-1}$ denoting the given historical data, and $\boldsymbol{X}_0$ represents the virtual placeholder time node that collects information about the whole sequence using zero initialization. As shown in Fig.~\ref{model}, the encoder handles temporal dependence and sensor dependence based on CAMs. Specifically, a CAM compresses MTS information into the virtual placeholder time node to update $\boldsymbol{X}_0$ to obtain the sequence embedding $\boldsymbol{X}_e$.

Moreover, the encoder consists of multiple layers of CAMs, which are stacked to improve the model capacity, and each CAM computes and transfers the corresponding $\boldsymbol{X}_e$ to the decoder. Since anomalies tend to have a low reconstruction probability when the model is trained on normal data only~\cite{OmniAnomaly}, $\boldsymbol{X}_e$ is further embedded into the low-dimensional space through a series of fully connected layers to improve the detection performance. Then, it is reconstructed through a series of opposite fully connected layers and transmitted to the decoder.

\subsection{Decoder}
As shown in Fig.~\ref{model}, to make efficient use of the time series data, two distinct decoders for reconstruction and prediction are created. Both decoders consist of temporal self-attention layers with a lower triangular form mask. In this way, self-attention is bidirectional in the encoder and unidirectional in both decoders.

Furthermore, both decoders have the same number of network layers as the encoder, and each layer in the decoder corresponds to a layer in the encoder. Each layer in the encoder transmits an embedding $\boldsymbol{X}_e$ to the decoders, and the corresponding layer of the decoders accepts the embedding $\boldsymbol{X}_e$ and concatenates it to the original sequence. The difference between the two decoders is that one is used to predict the next value of the temporal data, while the other is used to reconstruct the entire historical sequence.

In the prediction decoder, the input to the decoder $\boldsymbol{X}_{feed\_pre}$ is $\boldsymbol{X}_0$ because only one-step prediction is required. The last timestamp of the output sequence is a prediction of the next moment, which can observe the entire historical series. Therefore, we intercept the last timestamp and obtain the prediction result $\boldsymbol{Y}_{:, K+1}$ using a fully connected layer as follows:
\begin{equation}
  \boldsymbol{Y}_{:, K+1}=\text {FC}(\text {Cropping}(\boldsymbol{H}^{M}_{pre}, -1)),
\end{equation}
where $-1$ denotes cropping from the end to obtain a sequence of length 1, $H_{pre}^M$ indicates the output of the stacked multiple attention layers in the decoder, and $FC$ denotes a fully connected layer.

The prediction loss is calculated using the root mean square error (RMSE) as follows:
\begin{equation}
  \mathcal{L}_{pre}=\sqrt{\frac{1}{N} \sum_{n=1}^{N}\left(\boldsymbol{X}_{n, K+1}-\boldsymbol{Y}_{n, K+1}\right)^2},
\end{equation}
where $\boldsymbol{X}_{n, K+1}$ and $\boldsymbol{Y}_{n, K+1}$ are the actual and predicted values of sensor $n$ at timestamp $K+1$, respectively.

\par In the reconstruction decoder, a placeholder $\boldsymbol{X}_0$ is added to the original sequence to reconstruct the entire series. The input to the reconstructed decoder can be expressed as $\boldsymbol{X}_{feed\_rec}=\{\boldsymbol{X}_0, \boldsymbol{X}_{:-1,:}\}$. Since the attention matrix in the decoder has a lower triangular form mask, the input and output will be misaligned by one timestamp. After multiple attention layers, we obtain a sequence of length $K$ by cropping from the tail and transform it to obtain the reconstruction result $\boldsymbol{Y}_{:,:K}$ as follows:
\begin{equation}
  \boldsymbol{Y}_{:, :K}= \text {FC}(\text {Cropping}(\boldsymbol{H}^{M}_{rec}, -K)),
\end{equation}
where $-K$ denotes cropping from the end to obtain a sequence of length $K$, and $H_{rec}^M$ indicates the output of the stacked multiple attention layers in the decoder.

The RMSE was also selected to calculate the reconstruction loss. The equation is expressed as follows:
\begin{equation}
  \mathcal{L}_{rec}=\sqrt{\frac{1}{K N} \sum_{t=1}^{K} \sum_{n=1}^{N}\left(\boldsymbol{X}_{n, t}-\boldsymbol{Y}_{n, t}\right)^2},
\end{equation}
where $\boldsymbol{X}_{n, t}$ and $\boldsymbol{Y}_{n, t}$ are the actual and reconstructed values of sensor $n$ at timestamp $t$, respectively.

\subsection{Joint Optimization}
When representing MTS data, the prediction loss is mostly used to predict a point in time, while the reconstruction loss is primarily used to capture the distribution of the time period. The combination of the two can reduce the difficulty of model training, which better characterizes temporal data and captures hidden features between sensors. Therefore, in the CAN model, the loss function is defined as the weighted sum of the two optimization objectives as follows:
\begin{equation}
  \mathcal{L}=\phi \mathcal{L}_{pre}+\varphi \mathcal{L}_{rec},
  \label{loss}
\end{equation}
where $\phi $ and $\varphi $ denote the hyperparameters used to make a tradeoff between the prediction and reconstruction models, respectively, and $\phi +\varphi =1$. In general, the reconstruction task is considered to be easier than the prediction task. Thus, $\varphi $ is greater than $\phi $ during the first few training epochs, and then, $\phi $ becomes more significant than $\varphi $ during the last few training epochs. The corresponding training algorithm is described in Algorithm~\ref{A_model}.
\begin{algorithm}
	\caption{Training Process}
	\begin{algorithmic}[1]
		\Require Training set $\mathcal{X}$ of datasets, the number of attention module layers $M$, the batch size is $B$, and the window length is $K+1$;	
		\Ensure Well Trained model;
    \State Divide the dataset $\mathcal{X}$ into $\mathcal{W}$ according to the sliding window.
		\State Initialize embedding vectors $\boldsymbol{E}$ for all sensors.
        \Repeat
        \State Sample a batch ($\boldsymbol{X}\in \mathbb{R}^{B\times (K+1)\times N}$) from $\mathcal{W}$.
        \State Construct model inputs $\boldsymbol{X}_{feed\_en}$, $\boldsymbol{X}_{feed\_pre}$ and $\boldsymbol{X}_{feed\_rec}$.
        \State Add positional embedding into model inputs to obtain $\boldsymbol{H}^0$, $\boldsymbol{H}_{pre}^0$ and $\boldsymbol{H}_{rec}^0$.
        \State $\boldsymbol{A}^g \gets Relu(\boldsymbol{E}\times \boldsymbol{E}^T)$.

        \For{$i=1,2,\cdots , M$}
          \State // \textit{Encoder layer}
          \State $\boldsymbol{H}^{i} \gets \text{Self-Attention}(\boldsymbol{H}^{i-1})$
          \State Compute $\boldsymbol{A}^l$ by $\boldsymbol{H}^{i}$ according to Equation~\eqref{local1} and Equation~\eqref{local2}.

          \State $\boldsymbol{H}^i \gets \text{Global-localGraphConv}(\boldsymbol{H}^i, \boldsymbol{A}^g, \boldsymbol{A}^l)$.
          \State $\boldsymbol{X}_e^i \gets \boldsymbol{H}^i[:-1]$.
          \State // \textit{AE between encoder layer and decoder layers}
          \State $\hat{\boldsymbol{X}_e} \gets \text{AE}({\boldsymbol{X}_e}$).
          \State // \textit{Decoder layer for prediction}
          \State $\boldsymbol{H}_{pre}^{i-1} \gets \text{Concat}(\hat{\boldsymbol{X}_e}$, $\boldsymbol{H}_{pre}^{i-1})$.
          \State $\boldsymbol{H}_{pre}^{i} \gets \text{MaskSelf-Attention}(\boldsymbol{H}_{pre}^{i-1})$.
          \State // \textit{Decoder layer for reconstruction}
          \State $\boldsymbol{H}_{rec}^{i-1} \gets \text{Concat}(\hat{\boldsymbol{X}_e}$, $\boldsymbol{H}_{rec}^{i-1})$.
          \State $\boldsymbol{H}_{rec}^{i} \gets \text{MaskSelf-Attention}(\boldsymbol{H}_{rec}^{i-1})$.
        \EndFor

        \State // \textit{Obtain the prediction result $\boldsymbol{Y}_{:,K+1}$ and reconstruction result $\boldsymbol{Y}_{:,:K}$ through cropping and the fully connected layer}
        \State $\boldsymbol{Y}_{:,K+1} \gets \text{FC}(\text{Cropping}(\boldsymbol{H}_{pre}^{M},-1))$.
        \State $\boldsymbol{Y}_{:,:K} \gets \text{FC}(\text{Cropping}(\boldsymbol{H}_{rec}^{M},-K))$.
        \State Optimize the parameters by minimizing the loss function defined in Equation~\eqref{loss}.
        \Until{convergence}
	\end{algorithmic}
	\label{A_model}
\end{algorithm}
\subsection{Model Inference}

The prediction decoder is utilized to detect anomalies, and the reconstruction module is only used to assist in the training. The reconstruction-based method is more suitable for describing anomalies over the entire period. In point-in-time anomaly detection based on reconstruction, the target timestamp can be at a different position in the time window, and it becomes difficult to utilize multiple reconstruction results. In addition, the reconstruction-based model may neglect sudden perturbations to disrupt periodicity in a time series, especially when the values still follow the normal distribution~\cite{MTADGAN2020}. Therefore, only the predicted results of the timestamps are used for MTAD. Accordingly, the predicted deviation for each sensor can be calculated according to Eq. \ref{eqerr}:
\begin{equation}
\label{eqerr}
  \operatorname{Err}_{n}(t)=\left|\boldsymbol{X}_{n,t}-\boldsymbol{Y}_{n,t}\right|,
\end{equation}
where $\operatorname{Err}_{n}(t)$ denotes the deviation of sensor $n$ in the $t$ timestamp.

\par As different sensors may possess significantly different characteristics, their deviation values may have different scales. To prevent the overall anomaly score from being overly dominated by the deviation values produced by a particular sensor, the deviation values are normalized for each sensor. Normalization based on mean and interquartile range is used as defined in the following equation:
\begin{equation}
  \boldsymbol{S}_{n}(t)=\frac{\operatorname{Err}_{n}(t)-\mu_{n}}{IQR_{n}},
\end{equation}
where $\mu_{n}$ is the mean of $\operatorname{Err}_{n}$, $IQR_{n}$ is the interquartile range of $\operatorname{Err}_{n}$, and $\boldsymbol{S}_{n}(t)$ denotes the normalized deviation of sensor $n$ at the $t$ timestamp.

To calculate the anomaly score at the $t$ timestamp, the $K^s$ sensor aggregation with the largest deviation value is selected, as anomalies tend to affect only a small number of sensors, which is expressed by the following equation:

\begin{equation}
  \begin{aligned}
    idx_t =& argtopK^s(\boldsymbol{S}(t)),\\
    s_t =& \sum_{i\in idx_t} \boldsymbol{S}_i(t),
  \end{aligned}
  \label{score}
\end{equation}
where $idx_t$ is a set of $K^s$ sensors with large deviation values. The larger the anomaly score, the more likely it is to be an anomaly.
According to a threshold defined by methods, such as the dynamic error threshold~\cite{hundman2018detecting} and peaks-over-threshold~\cite{OmniAnomaly}, when the anomaly scores of a timestamp are greater than the threshold, the timestamp is inferred to be an anomaly.

\subsection{Complexity Analysis}
The time complexity of the main components of the proposed CAN model is analyzed. We skip the batch dimension and analyze the complexity involved in multivariate temporal data of temporal length $K$ with $N$ sensors. The hidden dimension size of the temporal self-attention layer is \{$d_t$, $d_k$\}, the hidden dimension size of the global-local graph convolutional layer is $d_s$, and the sensor embedding dimension size is $d_e$. The main complex computation of the temporal self-attention layer contains the computation of feature mapping of $Q$, $K$, $V$ and the attention coefficients. Their complexity is $O(K\times d_t \times d_k)$ and $O(K^2\times d_k)$ for each sensor, respectively. The temporal self-attention layer has a complexity of $O(N\times K^2 \times d_k)$. In addition, the global-local graph convolutional layer incurs $O(N^2(d_e+K\times d_s))$ time complexity. The computational complexity of the global graph is $O(N^2\times d_e)$, and the computational complexity of the local graph computation is $O(N^2\times K\times d_s)$. The remainder no longer involves high-complexity multiplication operations.

\section{Experiments}
In this section, first, the datasets and evaluation metrics are introduced. Second, we describe the experimental setup and the prepared MTS data, selected baselines, and configuration of the proposed model. Finally, dedicated analyses are performed to illustrate the effectiveness of the proposed model.

\subsection{Datasets}
Three real-world MTS datasets, namely, the Secure Water Treatment (SWaT) dataset~\cite{li2019mad}, Water Distribution (WADI) dataset~\cite{li2019mad} and Soil Moisture Active Passive satellite (SMAP) dataset~\cite{hundman2018detecting}, were used for the evaluation. Specifically, SWaT was collected from a water treatment testbed for cybersecurity, where the testbed was a fully operational scaled-down water treatment plant with a small footprint. The dataset consists of 11 days (7 days under normal operation and 4 days with attacks) of continuous operation, and every second of the testbed's physical properties (25 sensors and 26 actuators) was recorded. The transducer names define their roles, e.g., MV denotes a motorized valve, P denotes a pump, FIT denotes a flow meter, and LIT denotes a level transmitter. The dataset contains 946,719 samples, of which 496,800 were collected under normal operation, while 449,919 were collected with attack scenarios. More details are available on the website$\footnote{https://itrust.sutd.edu.sg/testbeds/secure-water-treatment-swat/}$. WADI is a natural extension of SWaT. WADI has a more complex composition and is equipped with more analytical equipment. It consists of 789,371 samples from 14 days of continuous normal operation and 172,801 samples from two days collected in attack scenarios. Similar to SWaT, WADI is a collection of all 103 sensor and actuator data during the data collection period. Compared with SWaT, WADI contains attacks on PLC and the network and simulated physical attacks (such as water leakage and malicious chemical injection). More details are available on the website$\footnote{https://itrust.sutd.edu.sg/testbeds/water-distribution-wadi/}$. The SMAP dataset is a public dataset published by NASA. The dataset is a record of telemetry data from sensors in individual spacecraft with a set of relevant telemetry values labeled by experts at NASA~\cite{hundman2018detecting}. Table~\ref{datasets} summarizes the three datasets.

\begin{table}[tbp]
  \caption{Dataset description and statistics.}
  \smallerspacecaption
  \footnotesize
  \centering
  \setlength\tabcolsep{2pt}
  \begin{tabular}{c|cccccc}
    \toprule
    Dataset & No. of & No. of & Training & Testing & Anomaly & Attack \\
    name & entities & features & set size & set size & ratio & Durations \\
    \midrule
    SWaT & 1 & 51 & 496,800 & 449,919 & 12.13\% & 2-25mins \\
    WADI & 1 & 118 & 789,371 & 172,801 & 5.85\% & 1.5-30mins \\
    SMAP & 55 & 25 & 135,183 & 427,617 & 13.13\% & -- \\
    \bottomrule
  \end{tabular}
  \label{datasets}
\end{table}

\subsection{Evaluation Metrics}
The precision, recall, and F1-score were used as the evaluation metrics, which are defined in the following equations:
\begin{equation}
  \text { Precision }=\frac{\mathrm{TP}}{\mathrm{TP}+\mathrm{FP}}, \\
\end{equation}
\begin{equation}
  \text { Recall }=\frac{\mathrm{TP}}{\mathrm{TP}+\mathrm{FN}}, \\
\end{equation}
\begin{equation}
  \mathrm{F1}= \frac{2 \times\text { Precision } \times \text { Recall }}{\text { Precision }+\text { Recall }},
\end{equation}
where TP denotes the truly detected anomalies, FP represents the falsely detected anomalies, TN denotes the truly detected normal samples, and FN represents the falsely detected normal samples. Moreover, the related metric scores may vary along with the anomaly score threshold. Hence, an optimal global threshold is defined based on a grid search.

In practice, an anomaly is often within a segment of consecutively observed anomalies marked by corresponding timestamps. A segment anomaly is correctly detected if any timestamp in the time segment of the marked anomaly is detected as an anomaly. In the proposed model, a point-adjust~\cite{point/adjust} strategy is used to transform the anomaly detection results. Specifically, for a continuous anomaly segment, if there is an observation in it that is correctly identified as an anomaly, then the anomaly segment has observations that are considered to be correctly detected by the model. If the ground truth is observed to be normal, no adjustments are made to the detection results.

\subsection{Experimental Setup}
\subsubsection{Data Processing}
Since the SWaT and WADI datasets have a huge sample size, to speed up the model training, the datasets were downsampled every ten seconds by taking the median value. Moreover, to improve the stability, the raw data were standardized using min-max normalization. Specifically, the maximum and minimum values of each sensor feature in the training set were calculated and then used to standardize the training set and testing set as follows:
\begin{equation}
  \tilde{x}^n=\frac{x^n-\min \boldsymbol{X}^n_{\text {train }}}{\max \boldsymbol{X}^n_{\text {train }}-\min \boldsymbol{X}^n_{\text {train }}},
\end{equation}
where $\min \boldsymbol{X}^n_{\text {train }}$ and $\max \boldsymbol{X}^n_{\text {train }}$ denote the minimum value and the maximum value of sensor $n$ in the training set, respectively.

\subsubsection{Baselines}
To highlight the performance of the proposed method, it is compared with several baselines, which are listed as follows:
\begin{enumerate}
  \item PCA~\cite{LI201463}: Principal component analysis obtains a low-dimensional projection by extracting the main feature components of the data. The anomaly score is the reconstruction error of this projection.
  \item KNN~\cite{KNN}: K-nearest neighbors uses the observation's distance to its $k$ nearest neighbors as the anomaly score.
  \item AE~\cite{AE}: The autoencoder reconstructs the observation using an encoder and decoder composed of MLPs. Anomalies are frequently difficult to recreate, and the reconstruction errors are used as anomaly scores to detect anomalies.
  \item IF~\cite{liu2008isolation}: Isolated forests use a binary search tree structure called iTree to isolate samples and then detect anomalies by using isolating sample points.
  \item DAGMM~\cite{zong2018deep}: The Deep Autoencoding Gaussian Model combines a deep autoencoder and a Gaussian mixture model to generate a low-dimensional representation of the samples and reconstruction results. The samples' reconstruction errors are considered anomaly scores.
  \item LSTM-VAE~\cite{LSTM-VAE}: LSTM-VAE replaces the feed-forward network in the VAE with LSTM and uses the reconstruction error as the anomaly score.
  \item OmniAnomaly~\cite{OmniAnomaly}: OmniAnomaly uses the VAE as the main structure, and a GRU is utilized to capture the complicated temporal correlation of multivariate observations. The anomaly scores are calculated based on the sample reconstruction errors.
  \item MTAD-GAN~\cite{MTADGAN2020}: MTAD-GAN learns the temporal dependencies and interdependencies between variables using the attention mechanism and then trains using a combination of reconstruction and prediction models. The anomaly scores are composed of the reconstruction error and prediction error.
  \item GDN~\cite{GDN}: The Graph Deviation Network learns the graphic relationship between sensors and performs single-step prediction using an attention mechanism; the prediction error is treated as the anomaly score.
  \item GTA~\cite{GTA}: GTA learns the relationship between sensors and combines graph convolution and a transformer to build a single-step time series prediction model; the prediction error is considered an anomaly score.
  \item DVGCRN~\cite{DVGCRN}: DVGCRN combines a probabilistic generative network with a variational graph convolutional recurrent network to model both spatial and temporal fine-grained correlations and considers both reconstruction-based and forecasting-based losses to optimize MTS representations.
\end{enumerate}

\subsubsection{Implementation Details}
The proposed method was implemented based on PyTorch 1.9.0 with CUDA 10.2. All experiments were conducted on an NVIDIA GeForce RTX 2080Ti. The model input was a historical time series with a window size of 5 (50 in SMAP) for single-step prediction and sequence reconstruction. The dimension size of the sensor embedding was set to 10 for SWaT and SMAP and 20 for WADI. The number of relations extracted $K^m$ was set to 10, and the number of sensors extracted $K^s$ was set to 2. The AE module was set up as a two-layer linear layer with the hidden layer dimension \{8, 4, 8\}. The encoder and two decoders both have three self-attention layers. For the multi-head attention mechanism, the number of heads was set to 8. The $\alpha$ and $\beta$ parameters were initialized to 0.2 and 0.8 during training and transformed to 0.8 and 0.2 after four epochs. The models were trained using the Adam optimizer with a learning rate of 1e-4, which decays with the number of training epochs. The early stopping strategy was also applied during training, and the patience was set to 5, which means that training stopped if there were five consecutive decreases in performance.

\par To implement the baseline methods, the number of neighbors was 5 in the KNN method, and the hidden layer dimension was \{64, 32, 32, 64\} in the AE method. DAGMM$\footnote{https://github.com/tnakae/DAGMM/}$, OmniAnomaly$\footnote{https://github.com/NetManAIOps/OmniAnomaly}$, MATD-GAT$\footnote{https://github.com/ML4ITS/mtad-gat-pytorch}$, GDN$\footnote{https://github.com/d-ailin/GDN}$ and DVGCRN$\footnote{https://github.com/SigmaLab01/DVGCRN}$ were implemented and reused in the evaluation. As the current implementation of GDN does not use a point adjustment policy, to ensure fairness, a point adjustment policy was added. Finally, for GTA, its paper does not provide the source code. However, it uses the same dataset as ours. Hence, its reported performance is directly used for comparison.

\begin{table*}[htp]
  \caption{Performance on SWaT, WADI and SMAP datasets.}
  \smallerspacecaption
  \centering
  \renewcommand\arraystretch{1.3}
  \setlength\tabcolsep{7pt}
  \begin{tabular}{cccccccccc}
    \toprule
    \multirow{2}{*}{Method}  &  & SWaT &  &  & WADI & & & SMAP &  \\
    \cmidrule{2-10}
     & Precision & Recall & F1-score & Precision & Recall & F1-score & Precision & Recall & F1-score\\
    \midrule
    {KNN} & 0.9994 & 0.6542 & 0.7908 & 0.4617 & 0.2813 & 0.3496 & 0.6471 & 0.899 & 0.7526
    \\
    {PCA} & 0.9200 & 0.6852 & 0.7854 & 0.8330 & 0.4811 & 0.6100 & 0.8432 & 0.6753 & 0.75
    \\
    {AE} & 0.8773 & 0.8116 & 0.8432 & 0.2475 & 0.3917 & 0.3033 & 0.7217 & 0.693 & 0.7071
    \\
    {IF} & 0.9347 & 0.7480 & 0.8310 & 0.5025 & 0.7843 & 0.6126 & 0.8807 & 0.5651 & 0.6885
    \\
    {DAGMM} & 0.8539 & 0.7653 & 0.8072 & 0.4113 & 0.6083 & 0.4908 & 0.6334 & 0.9984 & 0.7124
    \\
    {LSTM-VAE} & 0.9482 & 0.7938 & 0.8641 & 0.4834 & 0.8439 & 0.6148 & 0.7164 & 0.9875 & 0.7555
    \\
    {OmniAnomaly} & 0.9904 & 0.7493 & 0.8531 & 0.9713 & 0.4036 & 0.5702 & 0.8013 & 0.8606 & 0.8299
    \\
    {MTAD-GAN} & 0.9438 & 0.7808 & 0.8546 & 0.9846 & 0.2436 & 0.3905 & 0.8396 & 0.9994 & 0.9126
    \\
    {GDN} & 0.9766 & 0.8251 & 0.8945 & 0.8103 & 0.7217 & 0.7634 & 0.9768 & 0.5541 & 0.7071 \\
    {GTA} & 0.9483 & 0.8810 & {\ul 0.9134} & 0.8391 & 0.8361 & 0.8376 & 0.8911 & 0.9176 & 0.9041
    \\
    {DVGCRN} & 0.9548 & 0.8227 & 0.8838 & 0.8142 & 0.8648 & {\ul 0.8388} & 0.9730 & 0.9105 & {\ul 0.9407} \\
    {CAN} & 0.9133 & 0.9402 & \textbf{0.9266} & 0.8858 & 0.9026 & \textbf{0.8955} &0.9769 & 0.9111 & \textbf{0.9428}
    \\ \bottomrule
  \end{tabular}
  \label{overall}
\end{table*}

\subsection{Experimental Results}
Table~\ref{overall} reports the overall performance of the baselines and the proposed method on SWaT, WADI and SMAP. Since different methods have different threshold selection mechanisms, we tested the possible thresholds for each model and report the results with the highest F1-scores. On all datasets, CAN outperforms all previous techniques by achieving the best F1-score of $0.9266$ for SWaT, $0.8955$ for WADI and $0.9428$ for SMAP. On the SWaT dataset, many methods achieved high detection performance, but the best CAN F1-score still improves by approximately $2\%$ compared to the second-best method (GTA). While most approaches perform poorly on WADI, which has more complicated sensors, the best CAN F1-score is $6\%$ better than those of the second-best approach (DVGCRN). Additionally, because of the simpler sensor relationships in the SMAP dataset, the effect enhancement is less. On the precision index, the proposed model is marginally inferior to other techniques. Because we use the traversal strategy to obtain the threshold for all models, some models choose a lower threshold to increase precision and thus obtained a better F1-score. However, models with a low threshold, such as KNN and OmniAnomaly, have an extremely low recall. When considering the three indicators simultaneously, the effect of our approach remains ideal with balanced performance.

\begin{figure*}[htp]
	\centering  
	\subfloat[Dimension size of sensor embedding]{
		\includegraphics[width=0.32\linewidth]{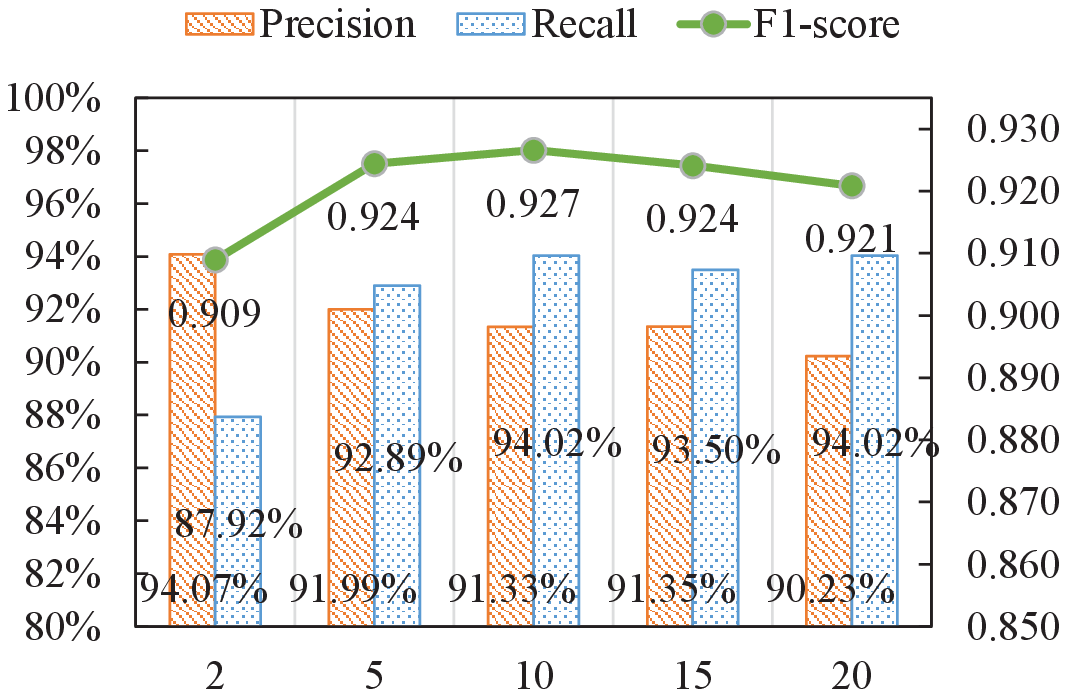}}
  \hfill
  \subfloat[Number of relations extracted]{
		\includegraphics[width=0.32\linewidth]{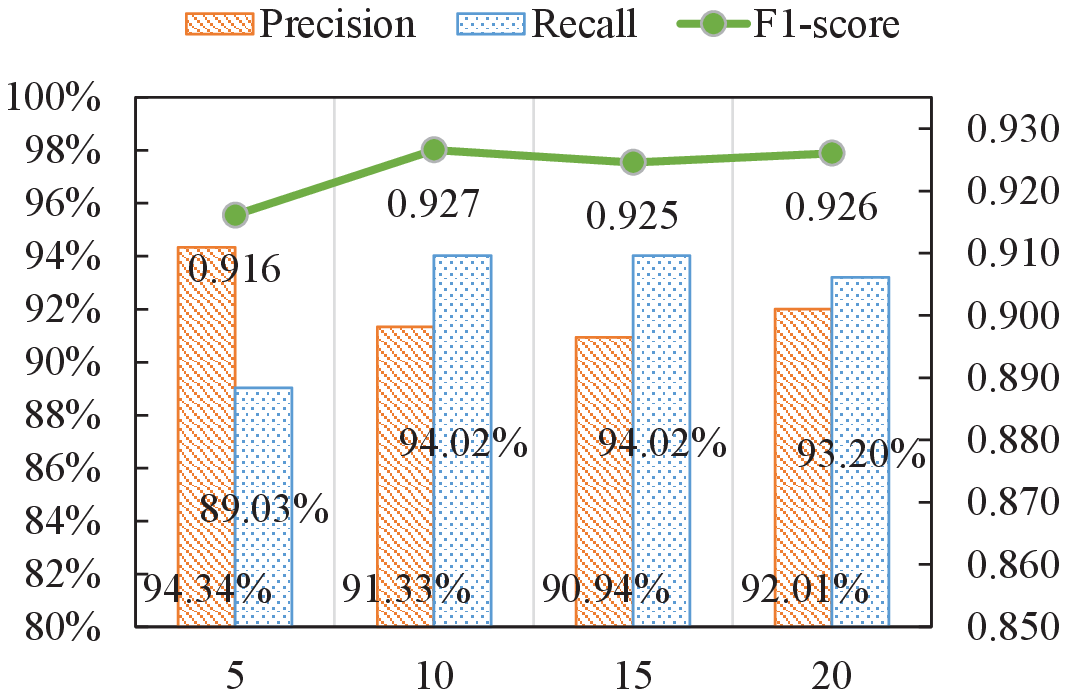}}
  \hfill
  \subfloat[Number of encoder (or decoder) layers]{
		\includegraphics[width=0.32\linewidth]{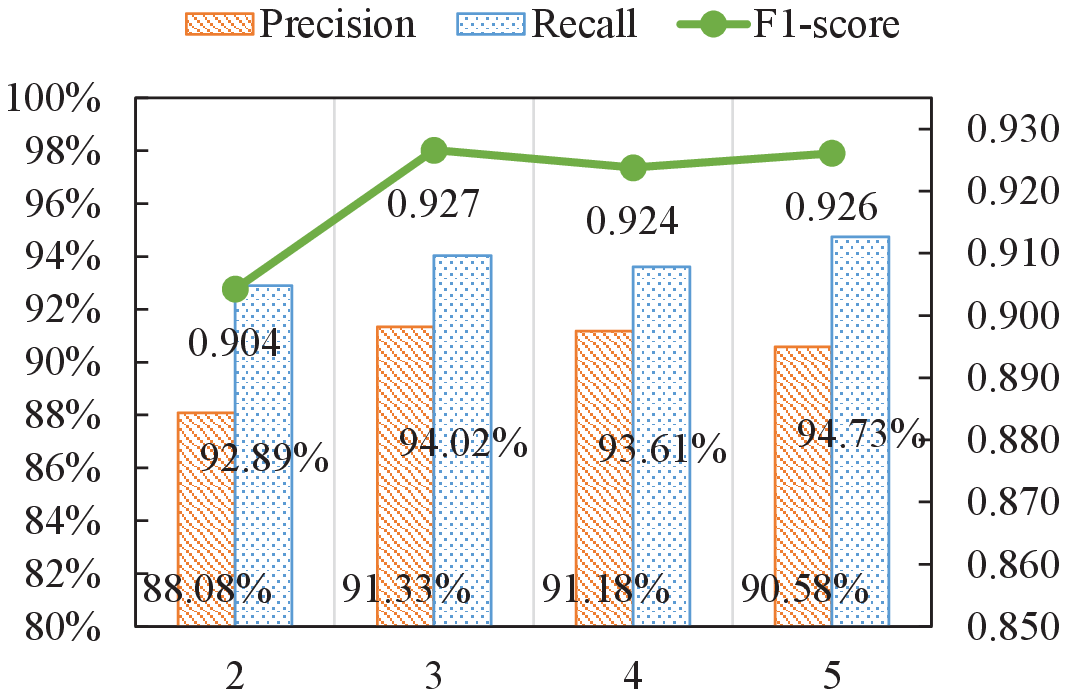}}
  \caption{Effect of parameters.}
  \label{param}
  \smallerspacecaption
\end{figure*}

The traditional methods, i.e., PCA, KNN, and AE, perform worse than deep learning-based methods. These machine learning methods are simple and fast, but their ability to process complex time series data is limited. There is no efficient way to model temporal correlation in multivariate time series data using traditional methods. To improve the performance, temporal dependencies should be properly measured. However, DAGMM directly detects each observation independently, completely ignoring the relationship between sensors. LSTM-VAE and OmniAnomaly accept multivariate continuous observations as input and use recurrent neural networks (LSTM, GRU) to model the temporal dependency. However, they do not explicitly learn the dependencies between sensors. Thus, they have less competitive performance than graph learning-based detection methods.
Specifically, MTAD-GAN uses an attention mechanism to model the dependencies between sensors, and it represents the relationships as a complete graph, which is unrealistic. GDN uses adaptive learning to learn the relationship between sensors but weakly models time series, leading to poor results on the SMAP dataset. GTA designs a directed graph structure learning approach to automatically learn the adjacency matrix among sensors, and within the model, dilated convolution and a transformer are used to learn the time dependence in time series. However, training such a complicated model is difficult. DVGCRN focuses on the robustness of the model under noise and does not model sensor correlations more accurately. CAN employs a global-local graph convolutional network to learn the correlation between variables in a comprehensive manner. CAN thus performs better when handling datasets that have complicated relationships. In addition, the proposed CAN method reduces the difficulty of model training by combining the reconstruction and prediction tasks.

\subsection{Parameter Sensitivity}
In this section, the effect of the CAN parameters is analyzed. Specifically, parameter sensitivity experiments were conducted on SWaT for the three parameters: dimension size of the sensor embedding, number of relations extracted, and number of encoder (or decoder) layers.

The sensor embedding records the global dependencies between sensors, and its dimension size is one key parameter. Fig.~\ref{param} shows the model performance results when the dimensions are 2, 5, 10, 15, and 20. The results show that the model performs optimally with a dimension size of 10. Too small of a dimension makes it difficult to include all the dependencies, while too large of a dimension makes it more challenging to learn the embedding vectors due to increased parameters. The number of extracted relations determines the number of dependencies established between sensors, which is crucial for learning dependencies between sensors. The most appropriate $K^m$ for SWaT is 10, which indicates that the dataset may have approximately ten dependencies for each sensor. Finally, parameter sensitivity experiments were performed on the number of layers in the model's encoder. The model performs poorly when the number of layers is 2, as it is difficult to simulate the complex dependencies in multivariate time series data. When the number of layers is increased to 3, the model's learning ability is improved, and the performance is significantly enhanced. However, when the number of layers continues to increase to 4 and 5, the model training difficulty increases, and the detection accuracy starts to decline slowly. In general, the model is robust in retaining a strong performance level across a wide range of parameters.

\subsection{Ablation Studies}
\begin{table}[tbp]
  \caption{Ablation experiments on the WADI dataset.}
  \smallerspacecaption
  \centering
  \begin{tabular}{@{}cccc@{}}
  \toprule
  Model elements & Precision & Recall & F1-score \\ \midrule
  w/o local graph & 0.8554  & 0.9115  & 0.8826 \\
  w/o graph convolution & 0.8364 & 0.8539 & 0.8451  \\
  w/o AE & 0.8436 & 0.9115 & 0.8539  \\
  w/o $\text{Decoder}_{rec}$ & 0.8413 & 0.9115 & 0.8750  \\
  $CAN^{+}$ & 0.8578 & 0.9115 & 0.8839 \\
  $CAN$ & 0.8885 & 0.9026 & \textbf{0.8955} \\ \bottomrule
  \end{tabular}
  \label{table:ablation}
\end{table}

To investigate the effectiveness of each component in CAN, we exclude the elements to observe how the related performance declines on WADI. Accordingly, five variant models of CAN were designed as follows:
\begin{enumerate}
  \item Without a local graph: The model only uses an adaptive static graph for graph convolution and removes the graph attention module.
  \item Without graph convolution: The graph convolution layer in CAN is replaced with a fully connected layer.
  \item Without AE: CAN without autoencoder between the encoder and decoders. Sequence embedding is transferred directly from the encoder to the prediction decoder and reconstruction decoder.
  \item Without $Decoder_{rec}$: The model is simplified to a prediction-based multivariate time series anomaly detection model.
  \item $CAN^+$: The method considers the reconstruction error when the target timestamp is the last timestamp of the input sequence to detect anomalies, and we fuse the reconstruction error and the prediction error in a certain proportion (0.1 is used in the paper) as the anomaly score.
\end{enumerate}

The results are summarized in Table~\ref{table:ablation}. Comparing the five variants, the complete model has the best performance. The performance degradation from the absence of both a local graph and graph convolution illustrates the necessity of inter-sensor dependencies for multivariate time series anomaly detection. The effectiveness without the local graph is reduced, demonstrating the local graph's significance in capturing the dependencies between sensors. The AE module has the effect of filtering some anomalies through reduction and reconstruction. A significant decrease in model effectiveness is observed when the model lacks the reconstruction decoder. This observation illustrates the effectiveness of the reconstruction decoder on multivariate time series data for encoder-assisted training. The reconstruction decoder helps the encoder represent multivariate time series data by reconstructing the task properly. In addition, the reconstruction error is typically influenced by the full input sequence and is not appropriate for timestamp anomaly detection directly; hence, the performance of $CAN^+$ is not improved in comparison with CAN.

\subsection{Visualization}
\subsubsection{Effect of the Global Graph}

\begin{figure}
	\centering
	\includegraphics[width=0.75\linewidth]{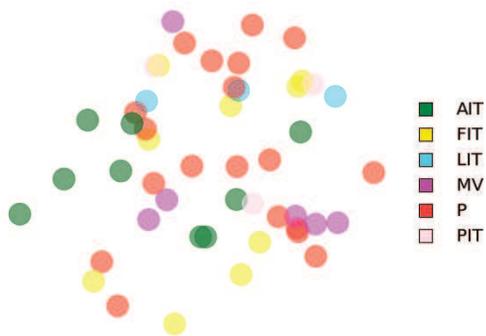}\\
	\makeatletter\def\@captype{figure}\makeatother
	\caption{t-SNE plot of the sensor embeddings.}
  \label{scatter}
  \smallerspacecaption
\end{figure}

To evaluate the effect of the model global graph, the sensor embedding vectors learned through training on SWaT are visualized based on t-SNE\cite{tSNE}, which is used to draw a scatter plot projecting the embedding vector into two-dimensional space. Accordingly, six different types of sensors with six different colors are presented in Fig.~\ref{scatter}. It can be observed that sensors of the same type tend to be close to each other. This observation also verifies that the learned embedding vectors can respond to a particular global feature of the sensors.

\subsubsection{Case Study}

\begin{figure}[htp]
	\centering  
	\subfloat[1\_MV\_001\_STATUS]{
		\includegraphics[width=0.48\linewidth]{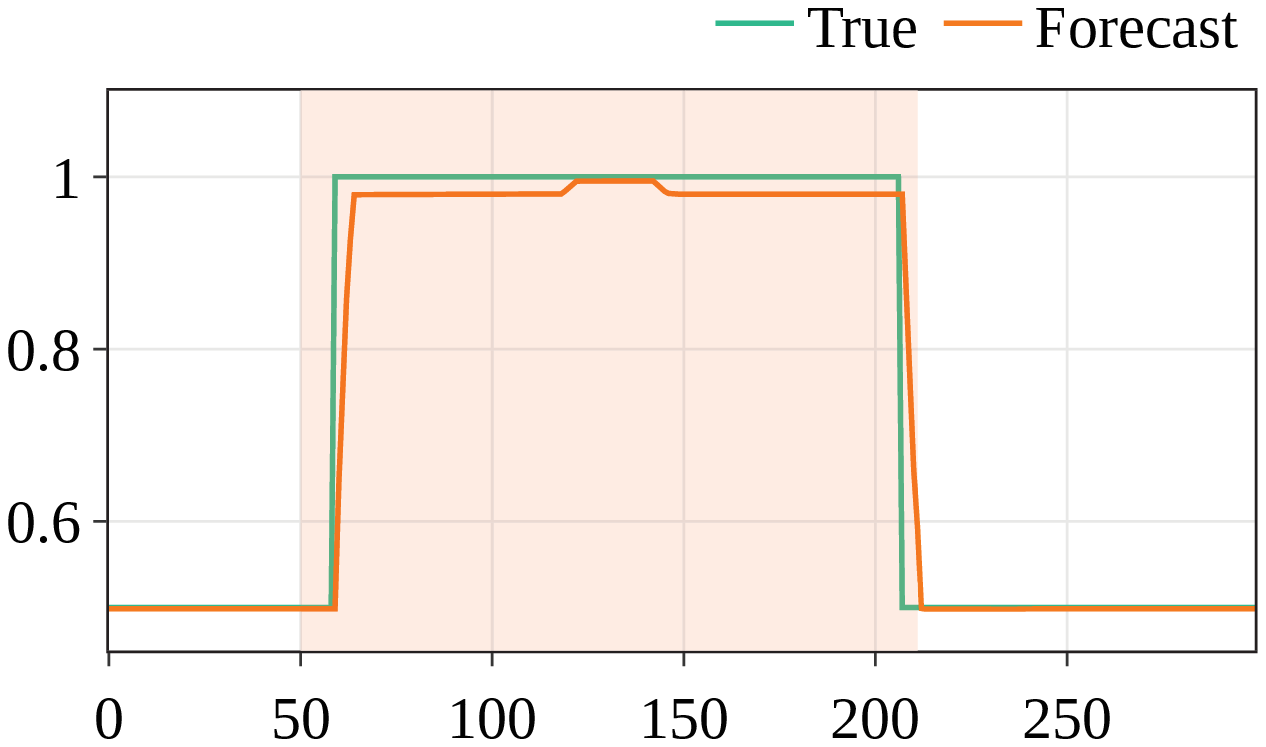}}
	\subfloat[1\_FIT\_001\_PV]{
		\includegraphics[width=0.48\linewidth]{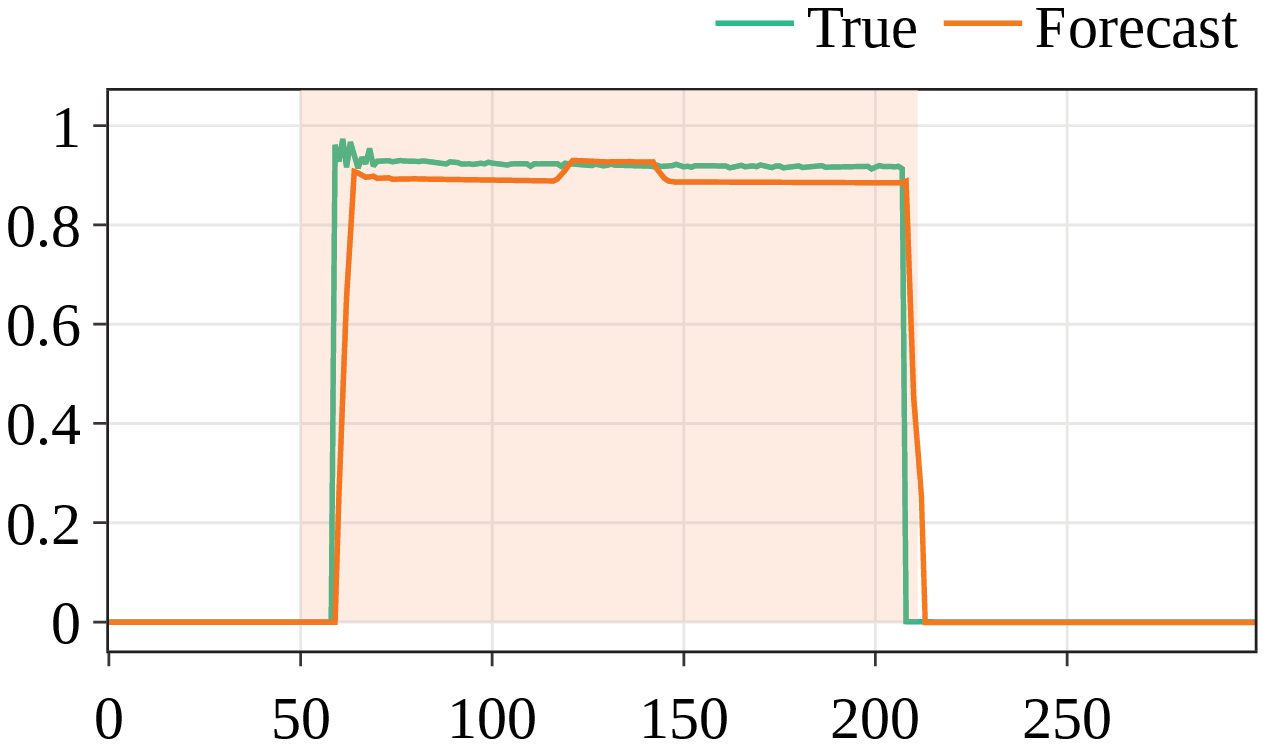}}
  \quad
	\subfloat[1\_LT\_001\_PV]{
		\includegraphics[width=0.48\linewidth]{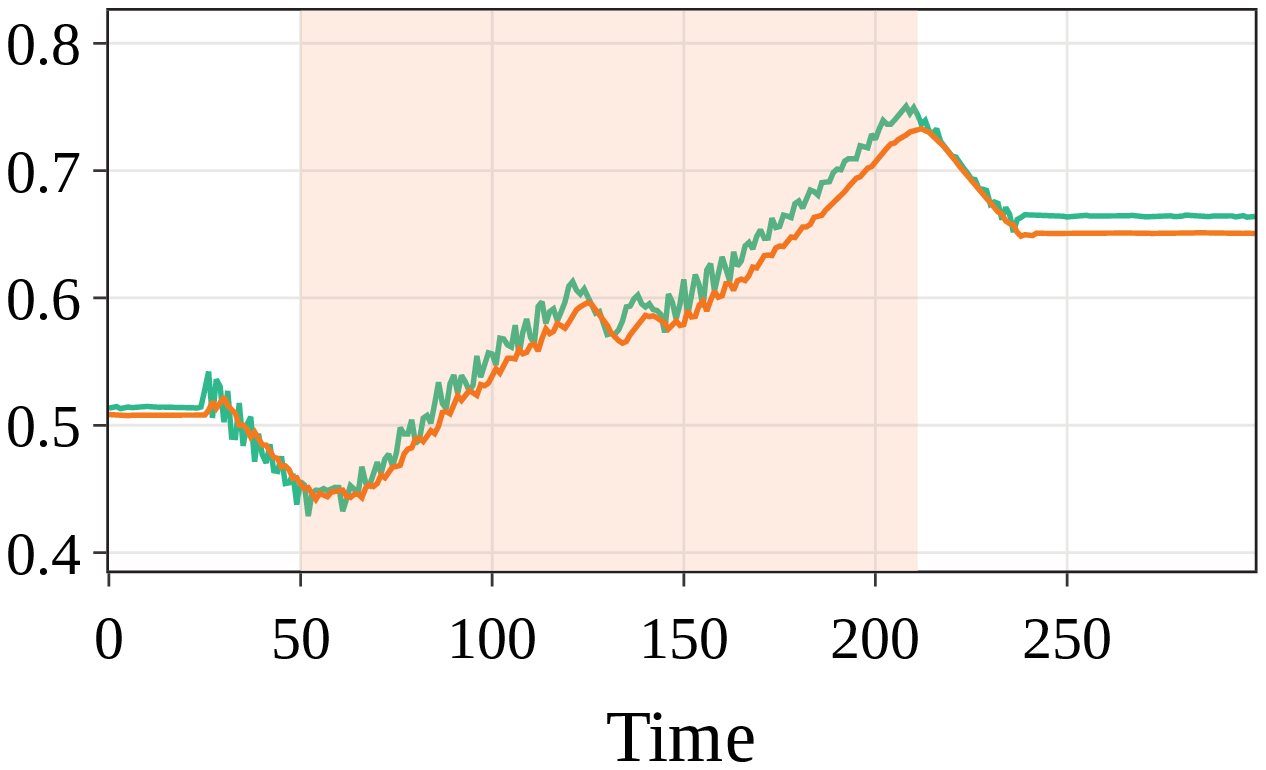}}
  \subfloat[2\_PIT\_001\_PV]{
		\includegraphics[width=0.48\linewidth]{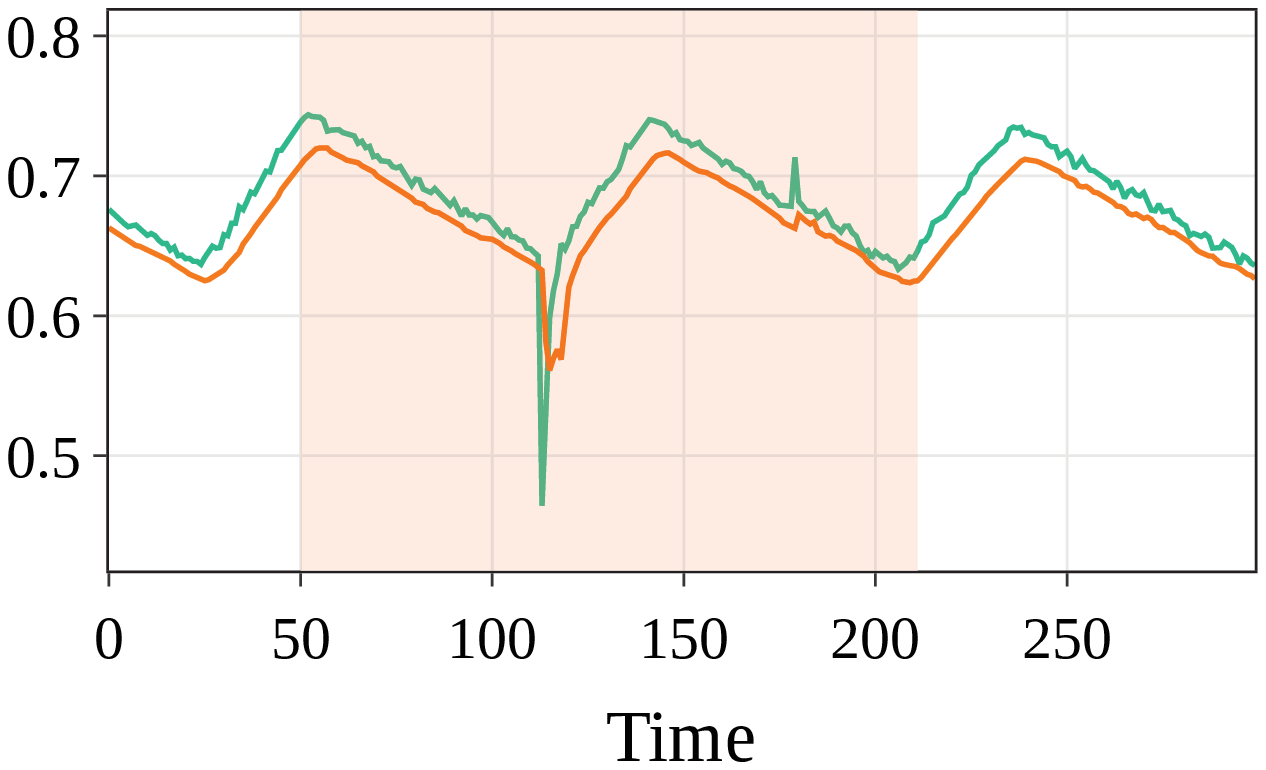}}
	\caption{The attacked sensor with three other sensors. The abscissa represents the timestamp, and the ordinate represents the sensor value.}
  \label{case}
  \smallerspacecaption
\end{figure}

A case study is prepared on an anomalous segment with a known attack in WADI to further evaluate our model. WADI's log file records an anomalous segment located between 19:25:00 and 19:50:16 on October 17. During this attack period, the attacker caused motorized valve 1\_MV\_001 to turn on maliciously to overflow the raw water tank.

The WADI test distribution treatment is divided into three stages: the water supply P1, distribution network P2, and return water system P3. P1 consists of two raw water tanks with a capacity of 2500 liters. Water intake into these two tanks can be from the water treatment plant named SWaT, from the public utility board inlet, or from the return water grid in WADI. Therefore, when the 1\_MV\_001 motorized valve is turned on, water flows into the raw water tank. The sensor 1\_FIT\_101 inflow meter increases rapidly, and the sensor 1\_LT\_101 level transmitter gradually climbs. At the same time, the attack on the motorized valve will also be transmitted to the P2 stage. For example, the pressure meter 2\_PIT\_001 sensor will change due to the increase in the P1 stage water level. In the global graph learned from training, 1\_MV\_001 and 1\_FIT\_101 are neighbors, and 1\_LT\_101 and 2\_PIT\_001 are neighbors to each other. The learned relationship is consistent with the actual relationship, which proves the effectiveness of the global graph we learned. Fig.~\ref{case} visualizes the transformation of the four sensors 1\_MV\_001, 1\_FIT\_101, 1\_LT\_101, and 2\_PIT\_001 in the anomaly segment; the green line represents the ground truth, the orange line represents the value calculated by the prediction model, and the red shading indicates the attack period. After 1\_MV\_001 was maliciously turned on, the values of 1\_MV\_001, 1\_FIT\_101, and 1\_LT\_101 were still in a normal range, but the 2\_PIT\_001 sensor in P2 was clearly out of the normal range and detected by the model. These results indicate that graph learning implemented in CAN is critical, and the proposed framework is effective and efficient in detecting multivariate time series anomalies.

\section{Conclusion}
In this work, we propose CAN, an anomaly detection framework based on self-attention and global-local graph convolution. It combines the advantages of predictive models and reconstruction models. In detail, to capture the dynamic inter-sensor dependencies in multivariate time series data, we propose to use a global graph to capture the static dependencies that depend on the features of the sensor itself and use graph attention to capture the dynamic sensor dependencies over time. Then, we combine temporal self-attention and graph convolution based on global and local graphs into a multilevel encoder-decoder framework. To better represent multivariate time series data, we jointly use reconstruction and prediction tasks to implement model optimization. When performing anomaly detection, we only compute anomaly scores that pass from the predictive model. A complex reconstructed model is prone to overfitting and ignores anomalies, and it is not suitable for anomaly detection at a single time point. Comprehensive experiments on three real datasets demonstrate that our model outperforms other state-of-the-art methods.
\par Most state-of-the-art methods use prediction-based methods for time series anomaly detection because of the overfitting problem in the reconstructed model. However, the reconstruction task is a crucial unsupervised task in time series data. In the representation learning of time series data, the reconstruction-based model has advantages that the prediction-based model does not have, such as robustness to perturbations and noise. This paper also demonstrates the effect of the reconstructed model on anomaly detection systems. However, the reconstruction task in our work only plays a role that is similar to pretraining. There must be a more appropriate method that can perfectly combine the predictive model and the reconstructed model for anomaly detection. Therefore, solving the overfitting of the reconstructed model and better combining prediction and reconstruction tasks will be a new direction for time series anomaly detection.

\bibliographystyle{IEEEtran}
\bibliography{references}

\begin{IEEEbiography}[{\includegraphics[width=1in,height=1.25in,clip,keepaspectratio]{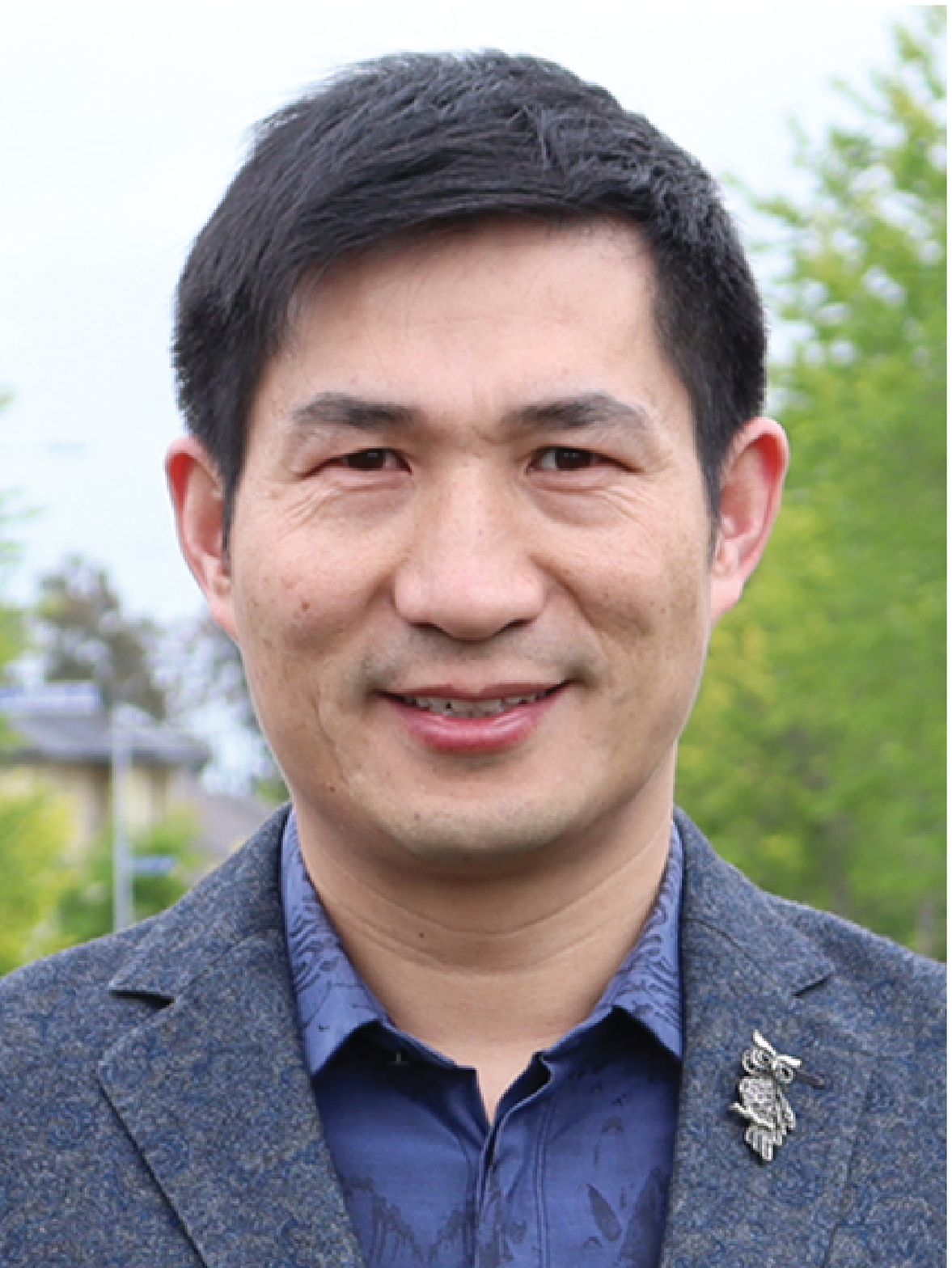}}]{Feng Xia} (Senior Member, IEEE) received BSc and Ph.D. degrees from Zhejiang University, Hangzhou, China. He is a Professor at the School of Computing Technologies, RMIT University, Australia. Dr. Xia has published 2 books and over 300 scientific papers in international journals and conferences (such as IEEE TAI, TKDE, TNNLS, TC, TMC, TPDS, TBD, TCSS, TNSE, TETCI, TETC, THMS, TVT, TITS, TASE, ACM TKDD, TIST, TWEB, TOMM, WWW, AAAI, SIGIR, WSDM, CIKM, JCDL, EMNLP, and INFOCOM). His research interests include data science, artificial intelligence, graph learning, and systems engineering. He is a Senior Member of IEEE and ACM and an ACM Distinguished Speaker.
\end{IEEEbiography}

\begin{IEEEbiography}[{\includegraphics[width=1in,height=1.25in,clip,keepaspectratio]{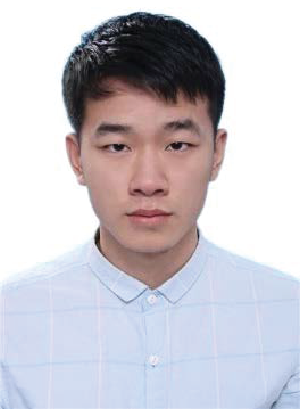}}]{Xin Chen} received a B.Sc. degree in information security from Harbin Engineering University, Harbin, China, in 2020. He is currently pursuing a master's degree from the School of Software, Dalian University of Technology, China. His research interests include spatiotemporal graph learning, urban science, and social computing.
\end{IEEEbiography}

\begin{IEEEbiography}[{\includegraphics[width=1in,height=1.25in,clip,keepaspectratio]{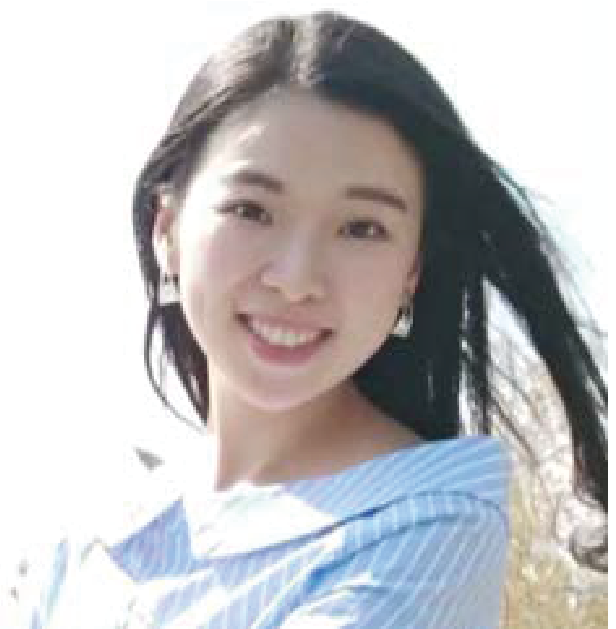}}]{Shuo Yu} (Member, IEEE) received B.Sc. and M.Sc. degrees from the School of Science, Shenyang University of Technology, China. She received a Ph.D. degree from the School of Software, Dalian University of Technology, China. Dr. Shuo Yu is currently an Associate Professor at the School of Computer Science and Technology, Dalian University of Technology. She has published over 50 papers and received several academic awards, including the IEEE DataCom 2017 Best Paper Award, IEEE CSDE 2020 Best Paper Award, and ACM/IEEE JCDL 2020 The Vannevar Bush Best Paper Honorable Mention. She has served as the Track Chair and PC member of several international conferences. Her research interests include data science, graph learning, and knowledge science.
\end{IEEEbiography}
\begin{IEEEbiography}[{\includegraphics[width=1in,height=1.25in,clip,keepaspectratio]{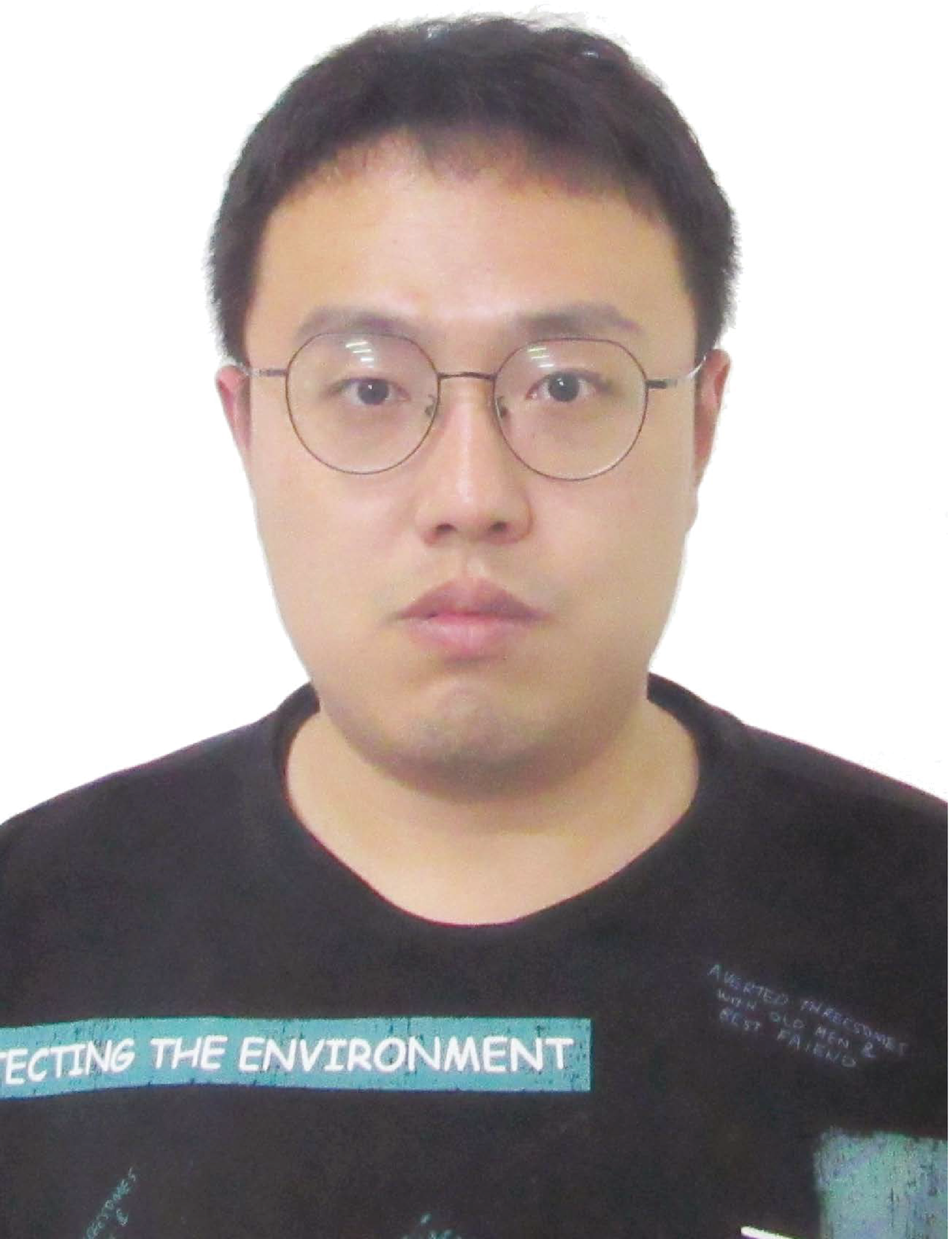}}]{Mingliang Hou} received a B.Sc. degree from Dezhou University and an M.Sc. degree from Shandong University, Shandong, China. He is currently pursuing a Ph.D. degree in software engineering with the Dalian University of Technology, Dalian, China. His research interests include graph learning, city science, and social computing.
\end{IEEEbiography}

\begin{IEEEbiography}[{\includegraphics[width=1in,height=1.25in,clip,keepaspectratio]{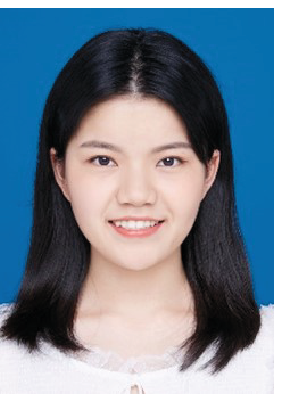}}]{Mujie Liu} received a B.Sc. degree from Ningbo Tech University, Ningbo, China, in 2021. She is currently pursuing a Ph.D. degree at the Institute of Innovation, Science and Sustainability, Federation University Australia, Ballarat, Australia. Her research interests include graph learning, anomaly detection, and artificial intelligence.
\end{IEEEbiography}
\begin{IEEEbiography}[{\includegraphics[width=1in,height=1.25in,clip,keepaspectratio]{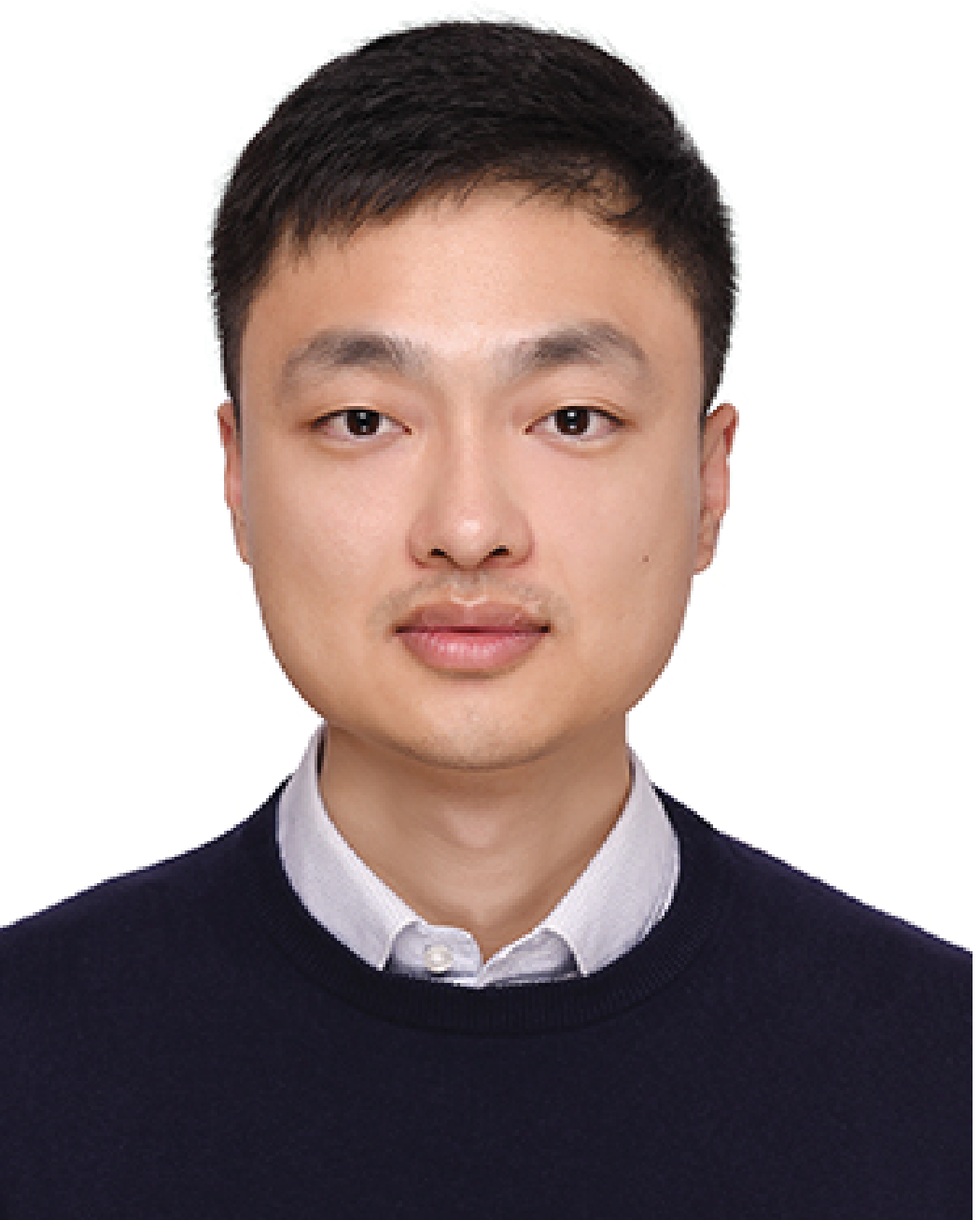}}]{Linlin You} (Member, IEEE) is an Associate Professor at the School of Intelligent Systems Engineering, Sun Yat-sen University, and a research affiliate at the Intelligent Transportation System Lab, Massachusetts Institute of Technology. He was a senior postdoc at the Singapore-MIT Alliance for Research and Technology and a research fellow at the Architecture and Sustainable Design Pillar of Singapore University of Technology and Design. He received his Ph.D. in Computer Science from the University of Pavia in 2015. He has published more than 40 journal and conference papers in the research fields of smart cities, service orchestration, multisource data fusion, machine learning, and federated learning.
\end{IEEEbiography}

\newpage

\end{document}